\documentclass{egpubl}
\usepackage{amsmath}
\usepackage{mathtools}
\usepackage{amssymb}
\usepackage{egsgp2022}
\usepackage{times}
\usepackage{bm}

\usepackage{comment}
\usepackage{booktabs}
\usepackage{amssymb}
\usepackage{color}
\usepackage{multirow}
\usepackage{wrapfig}
\usepackage{xcolor}
\usepackage{cite}
\usepackage[utf8]{inputenc}
\usepackage{mathtools}
\usepackage{comment}
\usepackage{booktabs} 
\usepackage{times}
\usepackage{epsfig}
\usepackage{xspace}
\usepackage{amssymb}
\usepackage{enumitem}
\usepackage{subfigure}
\usepackage{wrapfig}
\usepackage{verbatim}
\usepackage{caption}
\usepackage{etoolbox}
\usepackage{float}
\usepackage{calligra}
\usepackage{array}
\usepackage{float}
\usepackage{multicol}
\usepackage{amsfonts}
\usepackage{isomath} 
\DeclareMathOperator{\atan2}{atan2}

\newcommand{\eg}{\textit{e.g.}\xspace}
\newcommand{\etal}{\textit{et al.}\xspace}
\newcommand{\ie}{\textit{i}.\textit{e}.\xspace}

\DeclareMathOperator{\sign}{sign}

\newcommand{\surfit}{\textsc{SurFit}\xspace}
\usepackage{soul}

\SpecialIssueSubmission    

\usepackage[T1]{fontenc}
\usepackage{dfadobe}  

\usepackage{cite}  
\BibtexOrBiblatex
\electronicVersion
\PrintedOrElectronic

\usepackage{egweblnk}

\documentclass{egpubl}
\BibtexOrBiblatex
\usepackage{egsgp2022}
\usepackage{amsmath}
\usepackage{amsfonts}
\usepackage{mathtools}
\usepackage{amssymb}
\usepackage{times}
\usepackage{bm}
\usepackage{comment}
\usepackage{color}
\usepackage{multirow}
\usepackage{xcolor}
\usepackage{hyperref}
\hypersetup{colorlinks,citecolor=blue,filecolor=black,linkcolor=blue,urlcolor=blue} 
\usepackage[utf8]{inputenc}
\usepackage{booktabs} 
\usepackage{epsfig}
\usepackage{xspace}
\usepackage{enumitem}
\usepackage{subfigure}
\usepackage{wrapfig}
\usepackage{verbatim}
\usepackage{caption}
\usepackage{etoolbox}
\usepackage{calligra}
\usepackage{array}
\usepackage{float}
\usepackage{multicol}
\usepackage{isomath} 
\DeclareMathOperator{\atan2}{atan2}
\newcommand{\eg}{\textit{e.g.}\xspace}
\newcommand{\etal}{\textit{et al.}\xspace}
\newcommand{\ie}{\textit{i}.\textit{e}.\xspace}

\DeclareMathOperator{\sign}{sign}

\newcommand{\surfit}{\textsc{PriFit}\xspace}
\usepackage{soul}

\SpecialIssuePaper         
\usepackage[T1]{fontenc}
\usepackage{egweblnk}
\usepackage{dfadobe}  

\title[\textsc{PriFit}: Learning to Fit Primitives Improves Few Shot Point Cloud Segmentation]{\textsc{PriFit}: Learning to Fit Primitives Improves\\
Few Shot Point Cloud Segmentation}

\author[G.\ Sharma \etal]
{\parbox{\textwidth}{\centering G. Sharma$^{\dagger}$$^{1}$\orcid{https://orcid.org/0000-0002-7492-7808}, B. Dash\thanks{Equal contribution.}$^{1}$\orcid{https://orcid.org/0000-0002-8043-6669}, A. RoyChowdhury$^{1}$, M. Gadelha$^{1,2}$\orcid{0000-0002-4971-7980}, M. Loizou$^{3}$\orcid{0000-0002-2920-0087},  L. Cao$^{1}$, R. Wang$^{1}$,\\
E. G. Learned-Miller$^{1}$\orcid{0000-0002-3778-9135}, S. Maji$^{1}$\orcid{https://orcid.org/0000-0002-3869-9334} and E. Kalogerakis$^{1}$\orcid{https://orcid.org/0000-0002-5867-5735}}
        \\
{\parbox{\textwidth}{\centering $^1$University of Massachusetts Amherst  $^2$Adobe  $^3$University of Cyprus
       }
}
}

\begin{document}
\maketitle
\begin{abstract}
We present \surfit, a  semi-supervised approach for label-efficient learning of 3D point cloud segmentation networks. 
\surfit combines geometric primitive fitting with point-based representation learning. Its key idea is to learn point representations whose clustering reveals shape regions that can be approximated well by
basic geometric primitives, such as cuboids and ellipsoids. 
The learned point representations can then be re-used in existing network architectures for 3D point cloud segmentation,
and improves their performance in the few-shot setting. According to our experiments on the widely used ShapeNet and PartNet benchmarks,
\surfit outperforms several state-of-the-art methods in this setting, suggesting that decomposability into primitives is a useful prior for learning representations predictive of semantic parts.
We present a number of ablative experiments varying the choice of geometric primitives and downstream tasks to demonstrate the effectiveness of the method.

\end{abstract}  
\section{Introduction}
Several advances in visual recognition have become possible due to the supervised training of deep networks on massive collections of images. 
However, collecting manual supervision in 3D domains is  more challenging, especially for tasks requiring detailed surface annotations e.g., part labels.
To this end, we present \surfit, a \emph{semi-supervised} approach for learning 3D point-based representations, guided by the decomposition of 3D shape into geometric primitives. 

\begin{figure}[ht!]
    \centering
    \includegraphics[width=\linewidth]{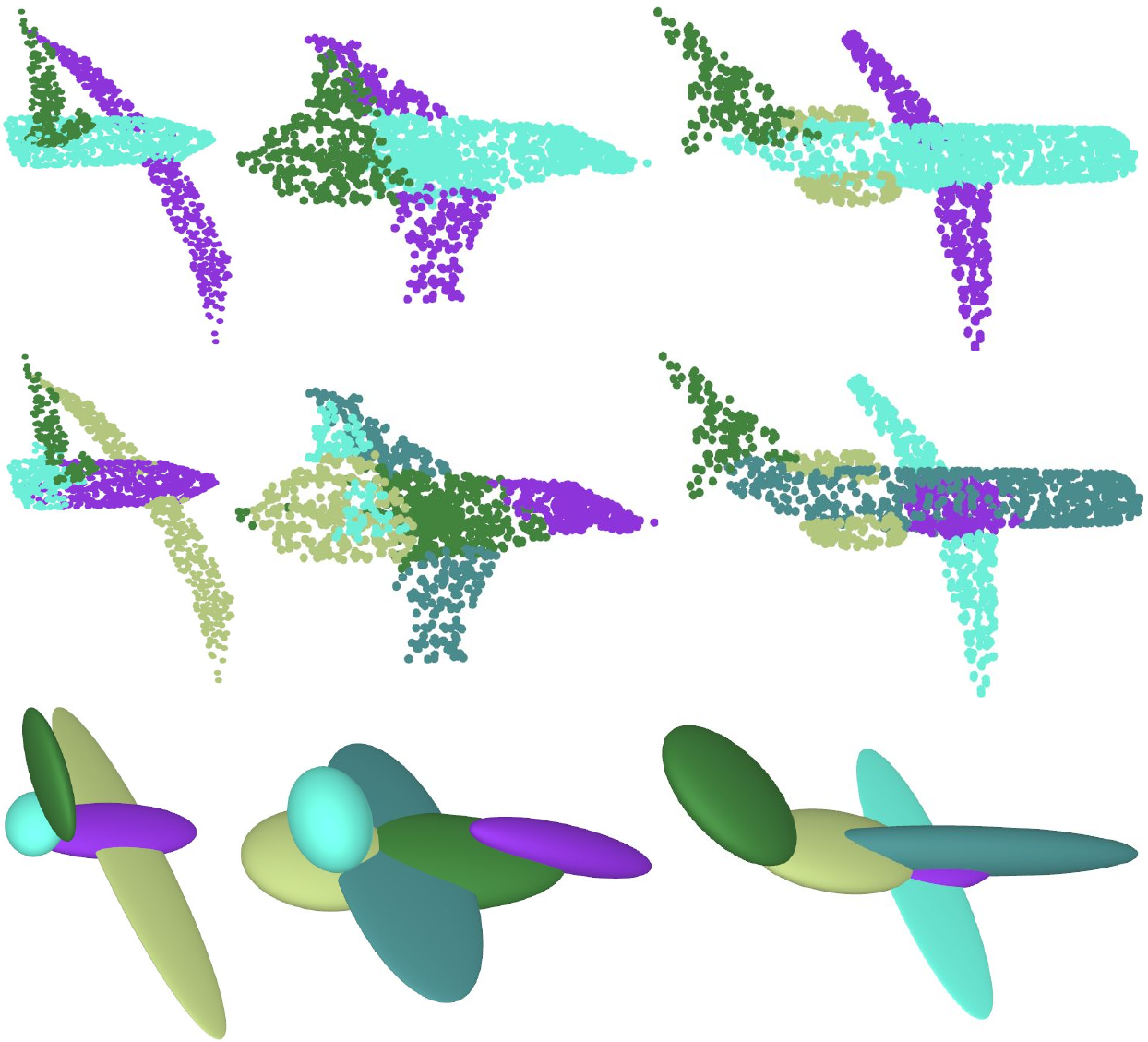}
   \caption{\textbf{\surfit uses primitive fitting within a semi-supervised learning framework to learn 3D shape representations.} \textit{Top row:} 3D shapes represented as point clouds, where the color indicates the parts such as wings and engines. The induced partitions and shape reconstruction obtained by fitting ellipsoids to each shape using our approach are shown in the \textit{middle row} and \textit{bottom row} respectively.
   The induced partitions often have a significant overlap with semantic parts.}\label{fig:splash}
\end{figure}

Our approach exploits the fact that parts of 3D shapes are often aligned with simple geometric primitives, such as ellipsoids and cuboids.
Even if these primitives capture 3D shapes at a rather coarse level, the induced partitions provide a strong prior for learning point representations useful for
part segmentation networks, as seen in Fig.~\ref{fig:splash}. 
This purely geometric task allows us to utilize vast amounts of unlabeled data in existing 3D shape repositories to guide representation learning for part segmentation. We show that the resulting representations are especially useful in the few-shot setting, where only a few labeled shapes are provided as supervision.  

\begin{figure*}[ht] \centering \includegraphics[width=\textwidth]{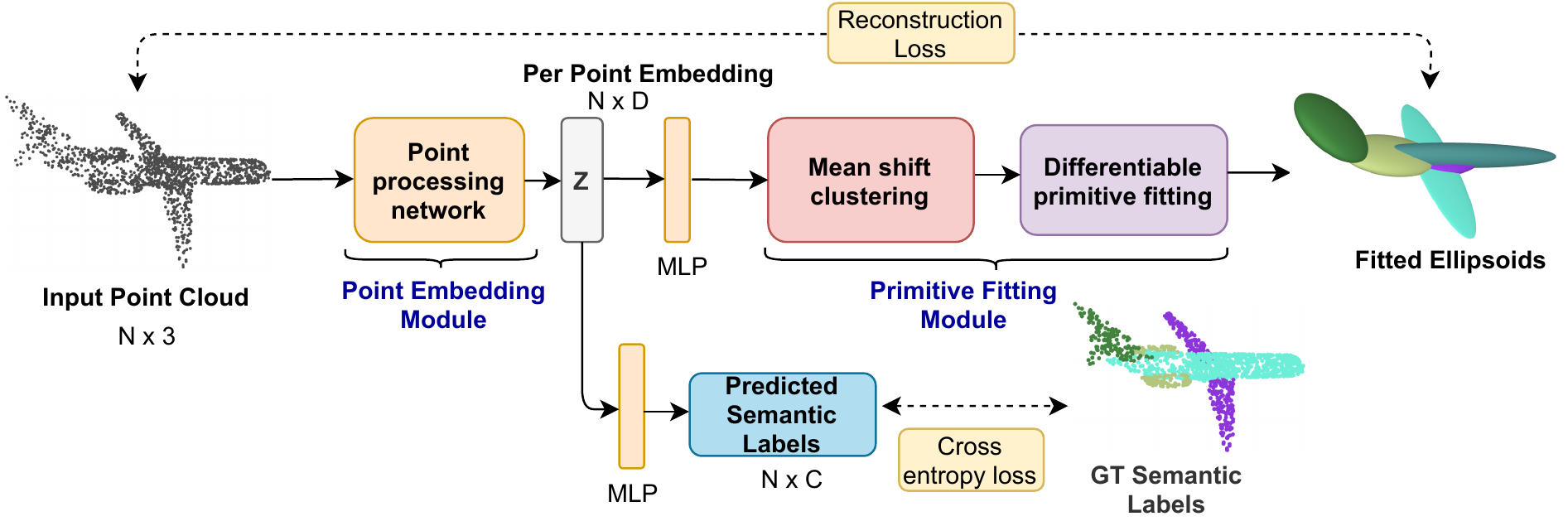}
 \caption{\textbf{Overview of \surfit}. Given a point cloud, the \textit{point-embedding module} outputs a feature representation for each point. This is processed through the \textit{primitive-fitting module}, that uses mean-shift clustering to cluster the points and fit a geometric primitive to each cluster.
    We train the network with a reconstruction loss between the fitted primitives 
    and the input point cloud over the unlabeled shapes, and a categorical cross 
    entropy loss over a small number of labeled shapes.}\label{fig:pipeline}
\end{figure*}

The overall framework for \surfit is based on a \emph{point embedding} module and a \emph{primitive fitting} module, as illustrated in Fig.~\ref{fig:pipeline}.
The point embedding module is a deep network that generates per-point embeddings for a 3D shape. Off-the-shelf networks can be used for this purpose (e.g., PointNet++~\cite{qi2017pointnetpp}, DGCNN~\cite{wang2019dynamic}). 
The primitive fitting module follows a novel iterative clustering and primitive parameter estimation scheme based on the obtained per-point embeddings. It is fully differentiable, thus, the whole architecture can be trained end-to-end.  
The objective is to minimize a reconstruction loss, computed as the Chamfer distance between the 3D surface and the collection of fitted primitives.
We experimented with various geometric primitives, including  ellipsoids or cuboids due to their simplicity. We further considered parameterized geometric patches
based on an Atlas~\cite{atlasnet}, as an alternative surface primitive representation.

Our method achieves $63.4$\%
part Intersection over Union (IoU) performance in ShapeNet segmentation dataset~\cite{Chang2015ShapeNetAI} with just one labeled example per-class, outperforming the prior state-of-the-art~\cite{selfsupacd} by $1.6\%$.
  We also present results on the PartNet dataset~\cite{partnet} where our approach provides $2.1$\% improvement compared to training from scratch approach while using $10$ labeled examples per-class. 
  We also present extensive analysis of the impact of various design choices, primitive types, size of the unlabeled dataset and different loss functions on the resulting shape segmentations.
Our experiments indicate that the use of ellipsoids as geometric primitives provide the best performance, followed by cuboids, then AtlasNet patches, as is shown in Table \ref{tab:sota_shapenet}.

\section{Related Work}
We are interested in learning per-point representations of 3D shapes in
a semi-supervised manner given a large number of unlabeled shapes and only a few labeled examples. 
To this end, we briefly review the literature on geometric primitive fitting and shape decomposition,
few-shot learning, and deep primitive fitting.
We also discuss the limitations of prior work and how we address them.

\paragraph*{Geometric primitives and shape decomposition.}
Biederman's recognition-by-components
theory~\cite{biederman1987recognition} attempts to explain object
recognition in humans by the ability to assemble basic shapes such as
cylinders and cones, called \textit{geons}, into the complex objects
encountered in the visual world. Early work in cognitive
science~\cite{hoffman1983parts} shows that humans are likely to
decompose a 3D shape along regions of maximum concavity, resulting in
parts that tend to be convex, often referred to as the
``\textit{minima rule}''. Classical approaches in computer vision
have modeled 3-D shapes as a composition of simpler
primitives, \eg work by
Binford~\cite{binford1987bayesian,binford1971visual} and
Marr~\cite{marr1978representation}.  More recent work in geometric
processing has developed shape decomposition techniques that generate
different types of primitives which are amenable to tasks like
editing, grasping, tracking and animation \cite{primsurveryKaiser}. Those have explored primitives like 3D curves~\cite{gsmc_iwires_sig_09,mzlsgm_abstraction_siga_09,Gori2017}, cages~\cite{obbcage}, sphere-meshes~\cite{spheremesh}, generalized cylinders~\cite{gdc}, radial basis functions~\cite{carr2001reconstruction,macedo2011hermite} and simple geometric primitives \cite{efficientransac}. This motivates the use of our geometric primitive fitting as a self-supervised task for learning representations.

\paragraph*{Unsupervised learning for 3D data.}
Several previous techniques have been proposed to learn 3D
representations without relying on extra annotations. Many such techniques rely on reconstruction
approaches~\cite{yang2018foldingnet,atlasnet,mrt18,zhao20193d,yang2019pointflow}. 
FoldingNet~\cite{yang2018foldingnet} uses an auto-encoder
trained with permutation invariant losses to reconstruct the point
cloud. Their decoder consists of a neural network representing a
surface parametrized on a 2D grid.
AtlasNet~\cite{atlasnet} proposes
using several such decoders that result in 
the reconstructed surface being
represented as a collection of surface patches. 
Li \etal~\cite{li2018sonet} presents SO-Net that models
spatial distribution of point cloud by constructing a self-organizing
map, which is used to extract hierarchical features. The proposed
architecture trained in auto-encoder fashion learns
representation useful for classification and segmentation.
Chen \etal \cite{chen2019bae} propose an auto-encoder (BAE-Net) with multiple branches, where each branch is used to reconstruct the shape by producing implicit fields instead of point clouds. However, this
requires one decoder for separate part, which restricts its use
to category-specific models. In contrast, our approach can train a single network in a category-agnostic manner
because our approach is based on inducing convexity priors in the embedding through primitive fitting -- this does not require category-specific knowledge, such as the number of semantic parts, making our approach more general.

Yang \etal \cite{Yangcuboidfitting} also propose fitting cuboids to point clouds for the task of co-segmentation. In order to segment the input point cloud based on the fitted cuboids, they define point to primitive membership based on point embeddings and enforce this membership to give correct correspondence between points and primitives. However, their method trains a category-specific model, thus suffers from the same limitation as BAE-Net.

Other techniques proposed
models for generating implicit functions from point
clouds~\cite{genova2019learning, deng2019cvxnet,occnet}, but it is
unclear how well the representations learned by those methods perform
in recognition tasks.
Several works use reconstruction losses along with other self
supervised tasks. Hassani \etal \cite{hassani2019unsupervised} propose multiple tasks:
reconstruction, clustering and classification to learn point
representation for shapes. Thabet \etal~\cite{MortonNet} propose a self-supervision
task of predicting the next point in a space filling curve (Morton-order
curve) using RNN. The output features from the RNN are used for semantic
segmentation tasks.  Several works have proposed learning point
representation using noisy labels and semantic tags available from
various shape repositories.  Sharma \etal \cite{SharmaPEN} learn point
representations using noisy part hierarchies and designer-labeled semantic tags
for few-shot semantic
segmentation. Muralikrishnan \etal~\cite{Muralikrishnan18} design a
U-Net to learn point representations that predicts user-prescribed
shape-level tags by first predicting intermediate semantic
segmentation. More recently, Xie \etal~\cite{PointContrast2020} learn
per-point representation for 3D scenes, where point embeddings of
matched points from two different views of a scene are pushed closer
than un-matched points under a contrastive learning framework.  Sun \etal \cite{NEURIPS2021_d1ee59e2} propose an approach for learning shape representations by canonicalizing point clouds with the help of capsule network while simultaneously decomposing point clouds into parts.

Closely related to our work, Gadelha et al.~\cite{selfsupacd} use approximate convex decomposition of  watertight meshes as source of self-supervision by training a metric over point clouds that respect the given decomposition.
Our approach directly operates on point clouds and integrates the decomposition objectives in a unified and end-to-end trainable manner. 
Empirically we observe that this improves performance. It also removes the need for having a black-box decomposition approach that is  separated from the network training.

\paragraph*{Semi-supervised learning for 3D data.}
Similar to our approach, Alliegro \etal \cite{jointssl}
use joint supervised and self-supervised learning for learning 3D shape features.
Their approach is based on solving 3D jigsaw puzzles as a self-supervised task to learn shape
representation for classification and part
segmentation. Wang \etal \cite{wang2020few} proposed semi-supervised approach that aligns a labeled template shape to
unlabeled target shapes to transfer labels using learned deformation. Luo \etal \cite{luo2020learning} proposed discovering 3D parts for objects in unseen categories by  extracting part-level features through encoding their local context, then  using agglomerative clustering and a grouping policy to merge small part proposals into bigger ones in a bottom-up fashion. Our method follows an orthogonal approach where features are learned and clustered into parts guided by primitive fitting.
\paragraph*{Deep primitive fitting.}
Several approaches have investigated the use of deep
learning models for shape decomposition.  Their common idea is to learn point-level representations used to generate
primitives. Several primitive types have been proposed,
including
superquadrics~\cite{superquadricshierarchical,superquadricsrevisited},
cuboids~\cite{shapehandles, abstractionTulsiani17}, generalized cylinder \cite{generalizedcylinderLiu} and radial basis
functions~\cite{genova2019learning}.
However, all these approaches have focused on \textit{generative}
tasks with the goal of editing or manipulating a 3D shape. 
Our insight is that reconstructing a shape by assembling simpler components improves representation learning for \textit{discriminative} tasks, especially when only a few labeled training examples are available.

\section{Method}
Our method assumes that one is provided with a small set of \emph{labeled} shapes
$\mathcal{X}_{l}$ and a large set of \textit{unlabeled} shapes $\mathcal{X}_u$.
Each shape $X \in \{\mathcal{X}_{l}, \mathcal{X}_u\}$ is represented as a point cloud with $N$ points, i.e., $X = \{x_i\}$ where $x_i \in \mathbb{R}^3$.
The shapes in $\mathcal{X}_l$ additionally come with part label $Y = \{y_i\}$ for each point.
In our experiments we use the entire set of shapes from the ShapeNet core dataset~\cite{Chang2015ShapeNetAI} and few labeled examples from the ShapeNet semantic segmentation dataset and PartNet dataset. 

The architecture of \surfit consists of a \emph{point 
embedding} module $\Phi$ and a 
\emph{primitive fitting} module ${\Psi}$.
The point embedding module ${\Phi}(X)$ maps the shape into embeddings corresponding to each point $\{{\Phi}(x_i)\} \in \mathbb{R}^D$.
The primitive fitting module ${\Psi}$ maps the set of point 
embeddings to a set of primitives $\{P_i\} \in {\mathcal{P}}$. Thus 
${\Psi} \circ {\Phi}: {X} \rightarrow {\mathcal{P}}$ is a 
mapping from point clouds to primitives.
In addition the point embeddings can be mapped to point labels via a classification function ${\Theta}$ and thus, ${\Theta} \circ {\Phi}: X \rightarrow Y$. 
We follow a \emph{joint training} approach where shapes from $\mathcal{X}_l$ are used to compute a supervised loss and the shapes from $\mathcal{X}_u$ are used to compute a self-supervised loss for learning by minimizing the following objective:

\begin{align}
 & \min_{{\Phi}, {\Psi},{\Theta}} {\mathcal{L}}_{\textsf{ssl}} + {\mathcal{L}}_{\textsf{sl}} \text{, where} \\
 {\mathcal{L}}_{\textsf{ssl}}  &= \mathop{\mathbb{E}}_{X \sim \mathcal{X}_u}  \Big[ \ell_{\textsf{ssl}} \big(X,{\Psi} \circ {\Phi}(X)\big) \Big]\text{, and} \\
 {\mathcal{L}}_{\textsf{sl}} &= \mathop{\mathbb{E}}_{(X,Y) \sim \mathcal{X}_l}  \Big[
    \ell_{\textsf{sl}}\big(Y,{\Theta} \circ {\Phi} (X)\big) \Big].\label{eq:jointraining}
\end{align}
Here $\ell_{\textsf{ssl}}$ is defined as a \textit{self-supervision loss} between the point cloud and a set of primitives, while $\ell_{\textsf{sl}}$ is the cross entropy loss between predicted and true labels. 
We describe the details of the point embedding module in Sec.\ref{sec:pem} and the primitive fitting module in Sec.~\ref{sec:decomposition}. 
Finally in Sec.~\ref{sec:training} we describe various loss functions used to train \surfit.

\subsection{Point embedding module}\label{sec:pem}
This module produces an embedding of each point in a point cloud.
While any point cloud architecture~\cite{qi2017pointnet,qi2017pointnetpp,wang2019dynamic}
can be used, we experiment with PointNet++
\cite{qi2017pointnetpp} and DGCNN \cite{wang2019dynamic}, two popular architectures for point cloud segmentation. These architectures have also been used in prior work on few-shot semantic segmentation making a comparison easier.

\subsection{Primitive fitting module\label{sec:decomposition}}
\noindent
The primitive fitting module is divided into a
\textit{decomposition} step that groups the set of points into clusters in the embedding space, and a \textit{fitting} step that estimates the parameters of the primitive for each
cluster.

\paragraph*{Decomposing a point cloud.}
These point embeddings ${\Phi}(x_i) \in \mathbb{R}^D$ are grouped into $M$ clusters using a differentiable mean-shift clustering. 
The motivation behind the choice of mean-shift over other clustering approaches such as k-means is that it allows the number of clusters to vary according to a kernel bandwidth.
In general we expect that different shapes require different number of clusters.
We use recurrent mean-shift updates in a differentiable manner as proposed by \cite{kong2018recurrent}. Specifically, we initialize seed points as ${G}^{(0)} = {Z} \in \mathbb{R}^{N \times D}$ and update them as follows:
\begin{equation}\label{eq:mean-shift}
{G}^{(t)} = {K} {Z} {D}^{-1}\text{.}
\end{equation}
We use the von Mises-Fisher kernel~\cite{mardia2009directional} $ {K}=\exp
({G}^{(t-1)} {Z}^T / b^2)$,
where ${D}=\texttt{diag}({K}\mathbb{I})$ and $b$ is the
bandwidth. The bandwidth is computed dynamically for each shape by using the average distance of each point to its $100^{th}$ neighbor in the embedding space~\cite{SilvermanDensity}.
${K}$ is updated after every iteration. 
The embeddings are normalized to unit norm, i.e., $||z_i||_2=1$, after each iteration. 
We perform $10$ iterations during training. 
After these updates, a non-max suppression step yields $M$
cluster centers $c_m, m=\{1,\ldots,M\}$ while making sure number of
clusters are bounded. The non-max suppression is done as follows: we start by extracting the highest density points that include at least one other point
within a prescribed radius $b$. Then we remove all points within that
radius and repeat until no other high density points are left. All
selected high density points in this way act as cluster centers.

Having updated the embeddings with the mean-shift iterations, we can
now define a \textit{soft membership} ${W}$ for each point
${x}_i$, represented by the embedding vector ${g}_i$\footnote{Superscript $t$ is dropped.}, to
the cluster center ${c}_m$ as
\begin{equation}\label{eq:membership}
    w_{i,m} =
    \frac{\exp({c}_m^\top{g}_i)}{\sum_{m}\exp({c}_m^\top{g}_i)} \text{,}
\end{equation}
where $w_{i, m}=1$ represents the full membership of the $i^{th}$
point to the $m^{th}$ cluster.

\paragraph*{Ellipsoid fitting.} Given the clustering, we then fit an ellipsoid to each of the clusters.  
Traditionally, fitting an ellipsoid to a point cloud is formulated
as a \textit{minimum volume enclosing ellipsoid}~\cite{mve} and solved using the Khachiyan algorithm. 
However, it involves an optimization procedure that is not readily differentiable, thus, making it hard to  incorporate in an end-to-end training pipeline. It is also susceptible to outliers (see supplementary
materials for details).
We instead rely on a simpler and faster differential approximation based on singular value decomposition (SVD) for ellipsoid fitting.

Given the membership of the point to $m^{th}$ cluster we first center the points and compute the SVD as

\begin{equation}\label{eq:mean}
    \mu = {W}_m {X} \mathcal{Z}^{-1}\text{,}
\end{equation}
\begin{equation}
    {\overline{X}} = {X} - \mu\text{, and}
\end{equation}
\begin{equation}\label{eq:svd}
    {U}, {S}, {V} =
    \texttt{SVD}({\overline{X}}^T{W}_m{\overline{X}}\mathcal{Z}^{-1})\text{.}
\end{equation}
Here ${W}_m$ is the diagonal matrix with $w_{i, m}$ as its
diagonal entries (${W}_m\left[i,i\right] = w_{i,m}$ for
$m^{th}$ cluster) and $\mathcal{Z} =
\texttt{trace}({W}_m)$. The orientation of an ellipsoid is given by
${V}$. 
The length of the principal axes can be computed from the singular values as $a_i = \kappa \sqrt{{S_{ii}}}$. We select $\kappa=\sqrt{3}/2$ by cross validation. The matrix $W_m$ selects the points with membership to the $m^{th}$ cluster in a `soft' or weighted fashion, and the SVD in Eq.~\ref{eq:svd} gives us the parameters of the ellipsoid that fits these weighted points.

\paragraph*{Discussion --- alternate choices for primitives.} Our approach can be used to fit cuboids instead of ellipsoids by considering the bounding box of the fitted ellipsoids instead. This may induce different partitions over the point clouds, and we empirically compare its performance in the Sec.~\ref{sec:exp}. 
Another choice is to represent the surface using an Atlas -- a collection of parameterized patches. 
We use the technique proposed in AtlasNet~\cite{atlasnet} where neural networks $f_\theta$ parameterize the coordinate charts $f_\theta:[0,1]^2 \rightarrow (x, y, z)$ conditioned on a latent code computed from the point cloud.

The decoders are
trained along with the encoder using gradient descent to minimize
Chamfer distance between input points and output points across all
decoders. 
An encoder trained in this fashion learns to decompose input points
into complex primitives, \ie via arbitrary deformations of the 2D
plane. 
Point representations learnt in this fashion by the encoder can
be used for downstream few-shot semantic segmentation task as shown in
Sec.~\ref{sec:exp} and Tab. ~\ref{tab:sota_shapenet}. However, this approach requires adding
multiple decoder neural networks.  Our \textit{ellipsoid fitting}
approach does not require significant architecture changes and
avoids the extra parameters required by AtlasNet.

\subsection{Loss functions \label{sec:training}}
\paragraph*{Reconstruction loss.}
This is computed as the Chamfer distance of input point clouds from predicted primitives. 
For the distance of a point on the input surface to the predicted surface we use
\begin{equation}\label{loss}
    \mathcal{L}_1 = \sum_{i=1}^N \min_{m} D_{m}^2({x}_i).
\end{equation}
Here $N$ is the number of points on the input shape and $D_m({x}_i)$ is the
distance of input point ${x}_i$ from the $m^{th}$
primitive. This is one side of Chamfer distance, ensuring that the
predicted primitives cover the input surface. 

In order to compute the  distance of a point $x$ from an ellipsoid or cuboid primitive, first the point is centered and re-oriented using the center and principal axes computed for the primitive
in Eq. \ref{eq:mean} and Eq. \ref{eq:svd} respectively: $p=V^T(x - \mu)$. Then we compute the signed distance $SD(p)$
of the point to each primitive, as described in the next paragraphs. Note that the unsigned distance used in Eq. \ref{loss} is simply computed as $D(p) = |SD(p)|$.
 
In the case of an ellipsoid primitive, we compute the approximate signed distance \cite{ellipsoidsdf} of the transformed point $p=(p_x,p_y,p_z)$ to it  as
\begin{equation}\label{eq:elllipsoidsdf}
    SD(p) = k_1 (k_1 - 1) / k_2 \text{,}
\end{equation}
where $k_{1}$ and $k_{2}$ are calculated as
\begin{equation}
    k_1 = \sqrt{(\frac{p_x}{s_x})^2 + (\frac{p_y}{s_y})^2 + (\frac{p_z}{s_z})^2}\text{ and}
\end{equation}
\begin{equation}
    k_2 = \sqrt{(\frac{p_x}{s_x^2})^2 + (\frac{p_y}{s_y^2})^2 + (\frac{p_z}{s_z^2})^2}
\end{equation}
Here $s=(s_x, s_y, s_z)$ is a vector storing the lengths of the ellipsoid semi-axis in all 3 directions.

In the case of a cuboid primitive, the distance of the transformed point $p$ to the primitive is computed 
as follows:
\begin{equation}
    q = \big(
   |p_x| - s_x, |p_y| - s_y, |p_z| - s_z)\text{,}
\end{equation}

\begin{equation}
    q_+ = (\max \{q_x, 0\}, \max \{q_y, 0\}, \max \{q_z, 0\})\text{, and}
\end{equation}
\begin{equation}\label{eq:cuboidsdf}
    SD(p) = ||q_+||_2 + \min\{\max\{q_x, q_y, q_z \}, 0\} \text{.}
\end{equation}
Here $s=(s_x, s_y, s_z)$ stores here the half-axes lengths of the cuboid. 
Note that the above equations measuring distances of points to primitives are analytic and differentiable with respect to the point coordinates.

To ensure that the input surface covers the predicted
primitives we minimize the loss
\begin{equation}
    \mathcal{L}_2 = \sum_{m=1}^{M} \sum_{{p} \sim E_m}
    \min_{i=1}^{N}||{x}_i - {p}||^2_2\text{,}
\end{equation}
where $E_m$ is the $m^{th}$ fitted primitive. We sample $10k$ points
over all ellipsoids, weighted by the surface area of each primitive. We
uniformly sample each primitive surface. These point samples are a function of the parameters of the predicted primitives and hence the gradients of the loss function can be back-propagated to the network. 
In the case of an ellipsoid, its parametric equation is:
\begin{align}
\label{eq:parameterization}
\begin{split}
\big( x, y, z \big) = \big(s_x\cos{u} \sin v, s_y \sin u \sin v, s_z\cos v \big),
\end{split}
\end{align}
and its inverse parameterization is:
\begin{align}\label{eq:inverse}
\begin{split}
        \big( v, u \big) = \big( \arccos(s_z / C), \atan2(s_y / B, s_x / A) \big), 
\end{split}
\end{align}
where ($u$, $v$) are the  parameters of the ellipsoid's 2D parametric domain, ($x$, $y$, $z$) the point coordinates, and ($s_x$, $s_y$, $s_z$) the lengths of semi-axis of the ellipsoid.
To sample the ellipsoid in a near-uniform manner, we start by creating a standard axis-aligned and origin centered mesh using the principal axis lengths predicted by \surfit. Then we apply Poisson surface sampling to gather points on the surface in an approximately uniform manner. We then compute 
parameters $(u, v)$ of the sampled points using Eq. \ref{eq:inverse}. 
Note that this is done outside the computation graph. We inject the 
computed parameters back to the computation graph using Eq. 
\ref{eq:parameterization} to compute point coordinates ($x,y,z$) again. 
These point coordinates are rotated
and shifted based on the predicted axis and center respectively. Sampling over the cuboid surfaces is done in a similar manner.

We use the two-sided loss to minimize reconstruction error
\begin{equation}\label{eq:total_loss}
    \mathcal{\ell}_{recon} = \mathcal{L}_1 + \mathcal{L}_2 \text{.}
\end{equation}

The hypothesis is that for a small number of primitives the
above losses encourage the predicted primitives to fit the input surface. 
Since the fitting is done using a union of convex primitives, each
diagonal entry of the matrix ${W}_m$ in Eq.~\ref{eq:svd} should have higher weights to sets of
points that belong to convex regions, thereby resulting in a convex
(or approximately~convex) segmentation of a point cloud. The point
representations learnt in this manner are helpful for point cloud segmentation as shown in Table~\ref{tab:sota_shapenet}.

\paragraph*{Intersection loss.} To encourage spatially compact clusters we introduce a loss function that penalizes overlap between ellipsoids.
Note that the clustering objective does not guarantee this as they operate on an abstract embedding space.
  Specifically, for each point ${p}$ sampled inside the surface of predicted shape should be contained inside a single primitive. 
  Alternatively the corresponding primitive should have negative signed distance $S_{m}({p})$ at that
  point ${p}$, whereas the signed distance (SD) should be positive for the remaining primitives.
  Let $\mathcal{V}_m$ be the set of points sampled inside the primitive $m$. Then intersection loss is defined as 
  \begin{equation}
      \ell_{inter} =\sum_{m} \sum_{p \sim \mathcal{V}_m} \sum_{j \neq m} \lfloor S_j({p})\rfloor_{-}^2,
  \end{equation}
  where $\lfloor S_m({p})\rfloor_{-} =~\min(S_m({p}), 0)$ includes only the points with negative SD, as points with positive SD are outside the primitive and do not contribute in intersection.
We use a differentiable approximation of the signed distances to
  ellipsoid and cuboid
  as described in Eq. \ref{eq:elllipsoidsdf} and Eq. \ref{eq:cuboidsdf}. 
  In the Table \ref{tab:intersection} we show that including intersection loss improves semantic segmentation performance.
  
\paragraph*{Similarity loss.}
Due to the general initialization schemes of the network weights, all the per-point embeddings are similar, which leads to mean-shift clustering grouping all points into a single cluster. This trivial clustering results in a single ellipsoid fitted to the entire shape which is similar to the PCA of the point cloud. This local minima results in negligible gradients being back-propagated to the network and prevents learning of useful features. We observed this phenomenon in several point-based network architectures.

This local minima can be avoided by spreading the point
embeddings across the space. 
We add a small penalty only at the early stage
of training to minimize the similarity of output point embeddings
${G}$ from mean-shift iterations (Eq.~\ref{eq:mean-shift}) as
\begin{equation}
\ell_{sym} = \sum_{i \neq j}(1 + {g}_i
{g}_j^\top)^2 \text{.}
\end{equation}
We discuss the quantitative effect of this loss in the Section \ref{sec:exp}.

\subsection{Training details}
We train our network jointly using both a self-supervised loss and
a supervised loss. We alternate between our self-supervised training while sampling point clouds from the entire unlabeled
$\mathcal{X}_u$ dataset and supervised training while taking
limited samples from the labeled $\mathcal{X}_l$ set. Hence our joint supervision and self-supervision approach follows semi-supervised learning paradigm.

 \begin{equation}
   L = \underbrace{\sum_{{X} \sim \mathcal{X}_u}
     \ell_{recon} + \lambda_1 \ell_{inter} +
     \lambda_2\ell_{sym}}_{\ell_{\textsf{ssl}} (\text{self-supervision})} +
   \underbrace{\sum_{{X} \sim \mathcal{X}_l}
     \mathcal{\ell}_{ce}}_{\ell_{\textsf{sl}} (\text{supervision})}
 \end{equation}

 where $\ell_{ce}$ is a cross entropy loss, $\lambda_1=0.001$ and
 $\lambda_2=2$ are constants. To produce segmentation labels for each point we implement the classification function ${\Theta}$ in Eq. \ref{eq:jointraining} using a 1D convolution layer followed by a softmax function. More implementation details are provided in the supplementary material.

\paragraph*{Back-propagation and numerical stability}
\begin{itemize}
\item To back-propagate the gradients through SVD computation we use
  analytic gradients derived by Ionescu \etal \cite{magnus99}. 
We
  implemented a custom Pytorch layer following \cite{Lingxiao}.
  SVD of a matrix $X\in \mathbb{R}^{m,n}$ is given by $X=U\Sigma V^T$ with $m\geq n$, $U^TU=I$, $V^TV=I$ and $\Sigma \in \mathbb{R}^{n,n}$ possessing diagonal structure. We define an SVD layer that receives a matrix X as input and produces a tuple of 3 matrices $U$, $\Sigma$ and $V$, defined as $f(X) = (U, \Sigma, V)$. Given a loss function $\mathcal L=L(f(X))$, we are interested in $\frac{\partial \mathcal L}{\partial X}$. Assuming, $\frac{ \partial \mathcal L}{\partial U},\frac{\partial \mathcal L}{\partial\Sigma},\frac{\partial \mathcal L}{\partial V}$ are known using automatic differentiation. For our purpose, we assume that $\frac{\mathcal \partial L}{\partial U}=0$ as $U$ is not a part of the computation graph.
   Then gradient of the output w.r.t.~the input $X$ is given by

\begin{equation}
    \frac{\partial \mathcal L}{\partial X} = U(\frac{\partial \mathcal L}{\partial\Sigma})_{diag}V^T + 2U\Sigma(K^T\circ(V^T(\frac{\partial \mathcal L}{\partial V})))_{sym}V^T,
\end{equation}
  \begin{equation}
      K_{ij} = \begin{cases}
    \frac{1}{\sigma_i^2 - \sigma^2_j},& i\neq j\\
    0,              & i=j.
\end{cases}
  \end{equation}
$\sigma_i$ is the $i^{th}$ singular value. $M_{sym} = \frac{1}{2}(M + M^T)$ and $M_{diag}$ is $M$ with off diagonal entries set to $0$.
   When
  back-propagating gradients through SVD, gradients can go to infinity
  when singular values are indistinct. This happens when membership
  weights in a cluster are concentrated on a line, point or sphere.
  Following \cite{Ionescu}, the above term $K_{ij}$ is changed to $K_{ij} = 1 / (\sigma_i +
    \sigma_j)\sign(\sigma_i -
    \sigma_j)(\max(|\sigma_i - \sigma_j|, \epsilon)$ with
  $\epsilon=10^{-6}$.
SVD computation can still be unstable when the condition number of the
input matrix is large.  In this case, we remove that cluster from the
backward pass when condition number is greater than $10^5$.
\item \textit{Membership function:} In Eq. \ref{eq:membership} the input
  to the exponential function is clamped 
  to avoid
  numerical instability during training. Furthermore, the quantity
  $r=\max_{i,m} {c}_m^T{g}_i$ is subtracted from the
  arguments of exponential function in both numerator and denominator
  to avoid weights becoming infinity.
\item \textit{Differentiability of mean shift clustering
  procedure:} The membership matrix ${W} \in
  \mathbb{R}^{N\times M}$ is constructed by doing non-max suppression
  (NMS) over output ${G}$ of mean shift
  iteration. The derivative of NMS w.r.t embeddings ${G}$ is
  either zero or undefined, thereby making it non-differentiable. Thus
  we remove NMS from the computation graph and back-propagate through
  the rest of the graph, which is differentiable through Eq.
  \ref{eq:membership}. This can be seen as a straight-through
  estimator \cite{Schulman2015}, which has been used in previous shape
  parsing works \cite{Lingxiao,sharma2020parsenet}.
\end{itemize}

\noindent The code of our implementation can be found on the web page: \href{https://hippogriff.github.io/prifit/}{https://hippogriff.github.io/prifit/}.

\section{Experiments\label{sec:exp}}
\begin{figure*}[]
    \centering
    \includegraphics[width=0.8\textwidth]{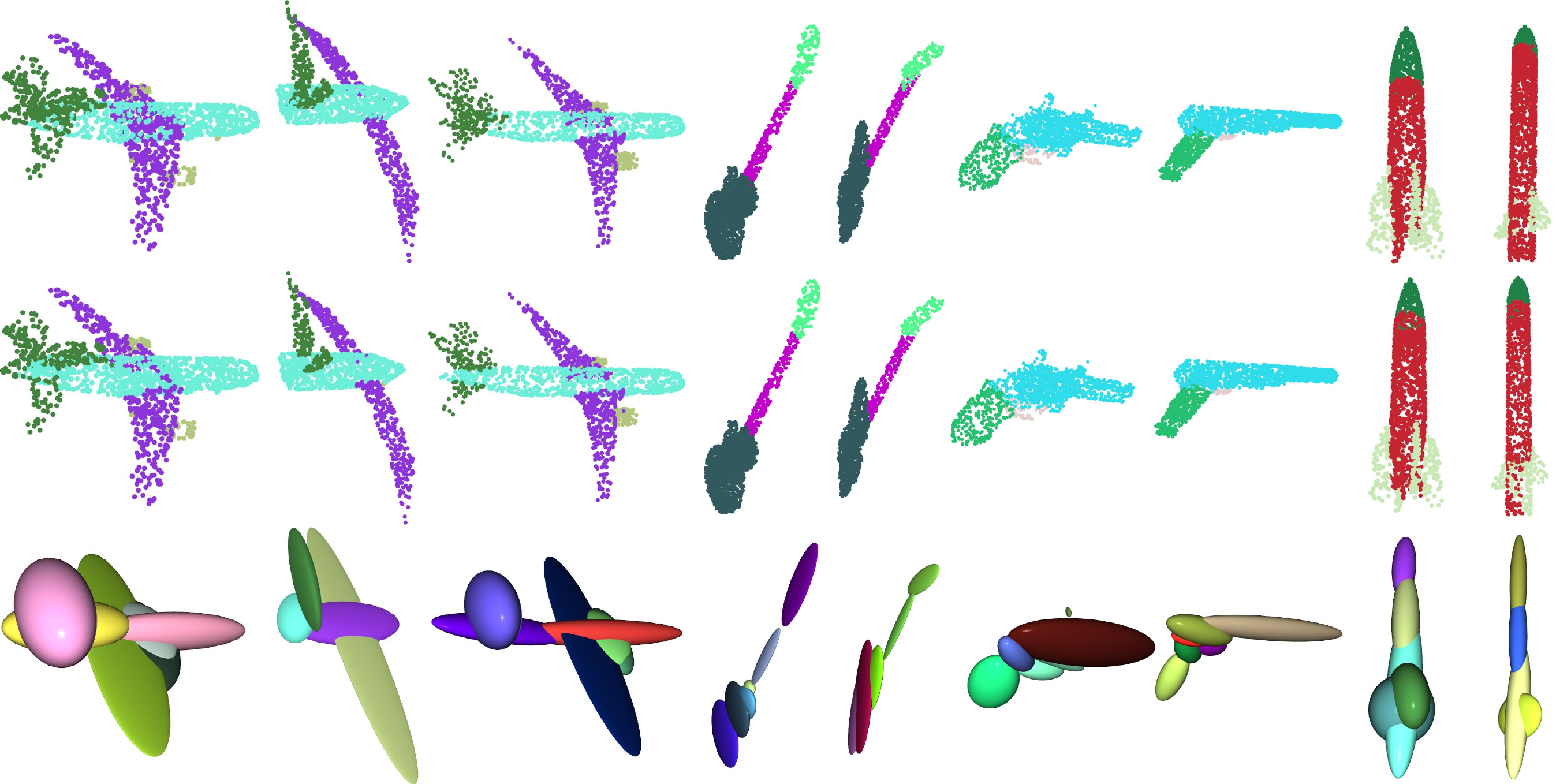}

   \caption{\textbf{Visualization of predicted semantic labels and
       ellipsoids on the ShapeNet dataset.} \textbf{Top:} Ground truth
     point clouds, \textbf{middle:} predicted labels using our fitting
     approach, trained using $k=10$ labeled examples per category,
     \textbf{bottom:} predicted ellipsoids. \surfit predicts variable number of ellipsoids to approximate the input point cloud while
     maintaining correspondence with semantic
     parts.}\label{fig:shapenet}
\end{figure*}

\begin{figure}[]
    \centering
    \includegraphics[scale=0.5]{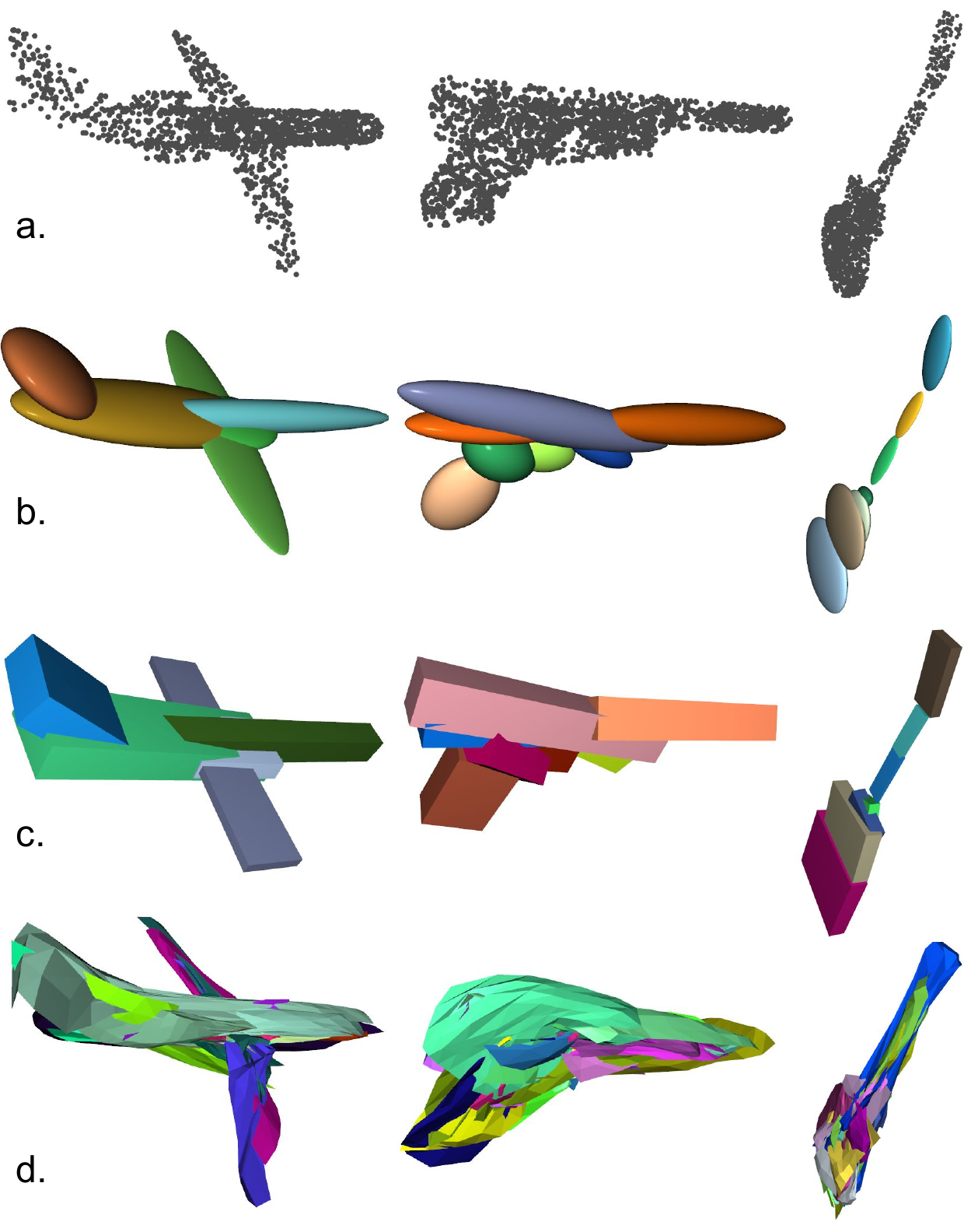}

   \caption{\textbf{Visualization of various primitive fitting
       approaches.} \textbf{a)} input point cloud. \textbf{b)} Ellipsoid fitting using our approach. \textbf{c)} cuboid fitting using our approach. \textbf{d)} different
     primitives from AtlasNet. Different colors are used to depict
     different primitives. For AtlasNet we visualize each chart with a
     unique color. Notice that geometric primitives are better
     localized and approximate the shape in fewer primitives in comparison
     to AtlasNet.}\label{fig:primitives}
\end{figure}

\begin{figure}
  \centering
\begin{tabular}{l}
  \includegraphics[width=0.8\linewidth]{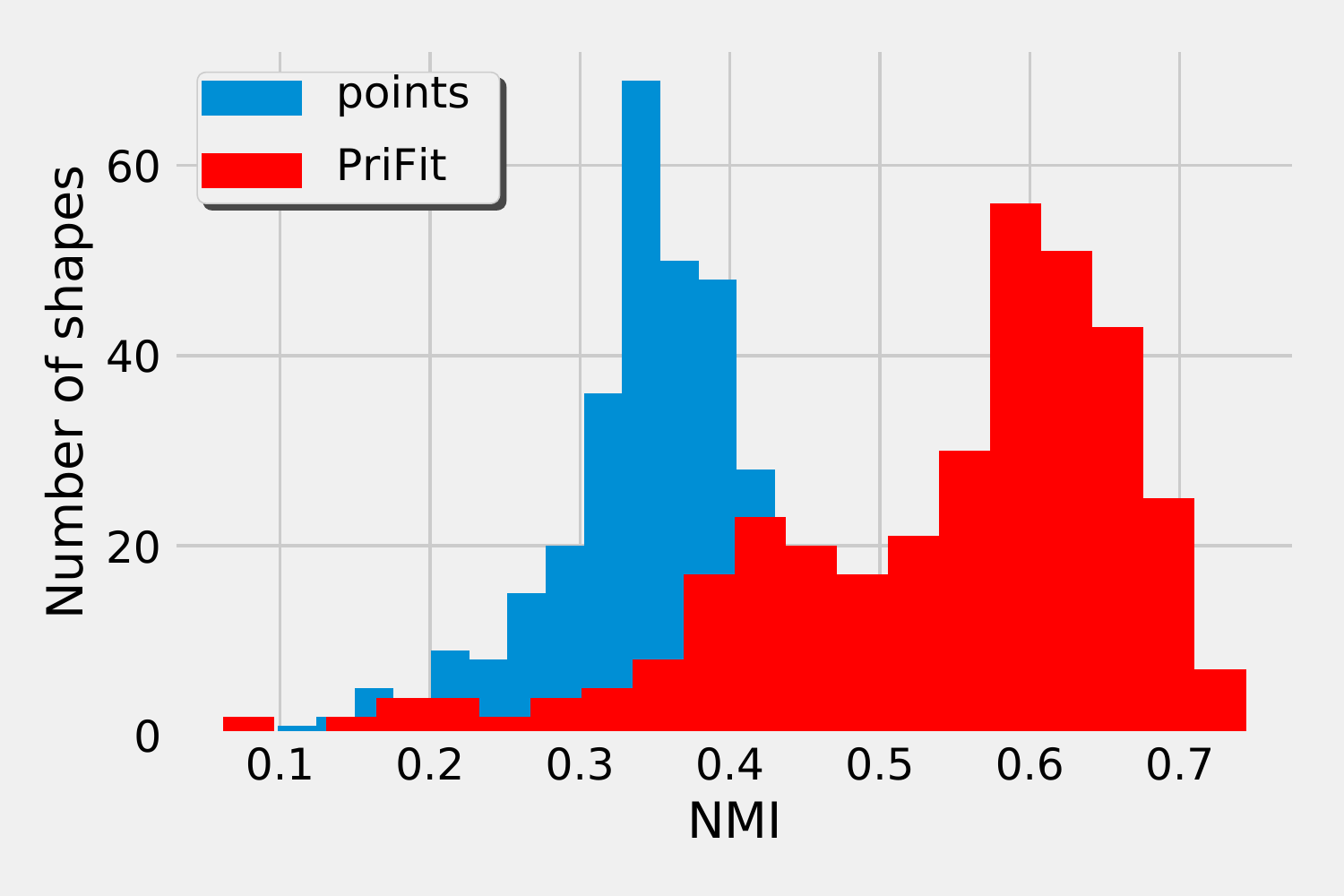}\\
  \includegraphics[width=0.8\linewidth]{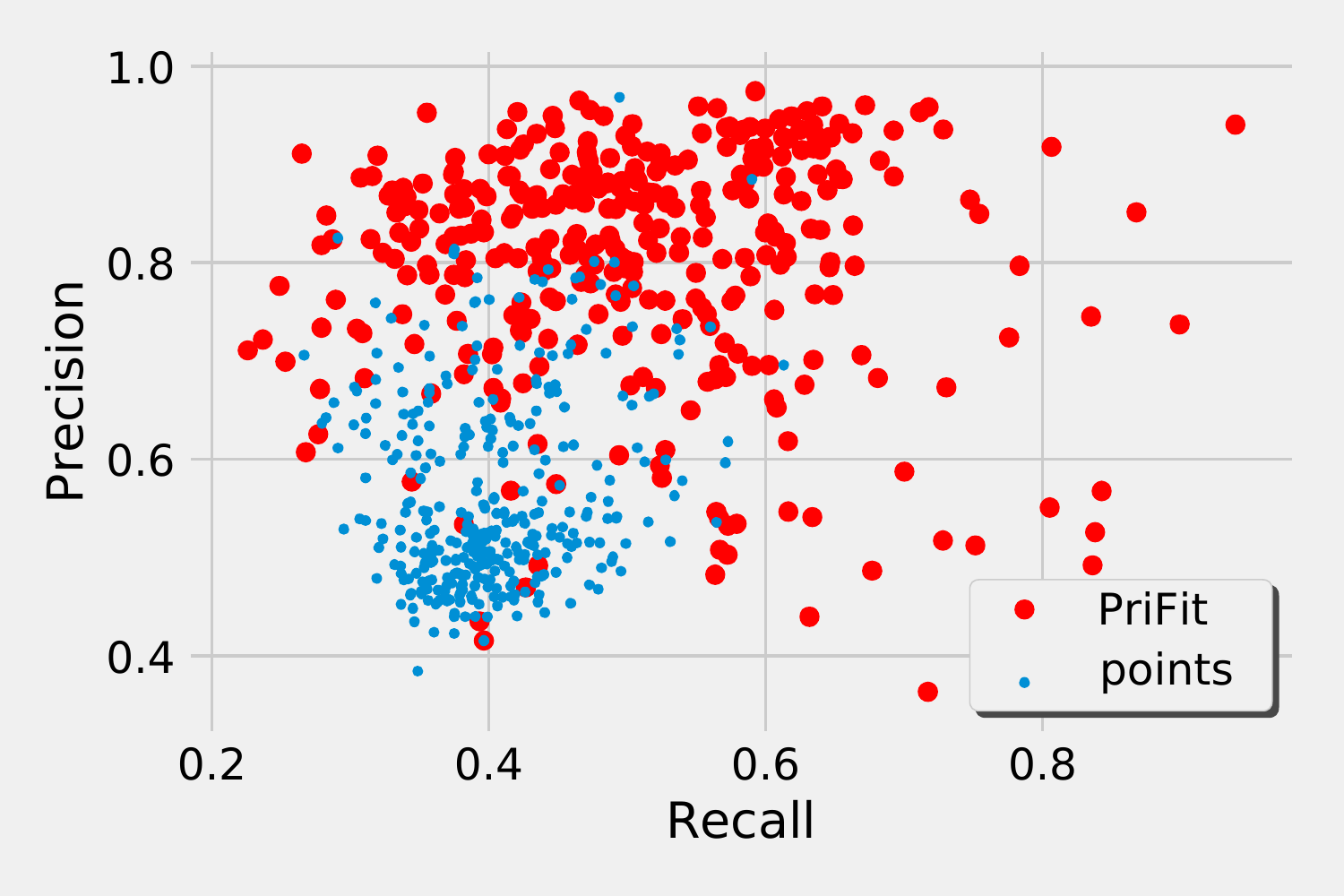}
\end{tabular}
\caption{\textbf{Analysis of clustering.}  We analyze two clustering
  approaches, 1) \surfit and 2) directly clustering
  \textit{points} using K-Means. \textit{Top:} normalized mutual information
  (NMI) and \textit{bottom:} precision vs recall  between predicted clusters
  and semantic part labels. \surfit gives higher average NMI ($54.3$ vs
  $35.4$) and higher precision than clustering with only points as
  features.}
  \label{fig:nmi}

\end{figure}

\begin{figure*}[]
    \centering
    \includegraphics[width=\textwidth]{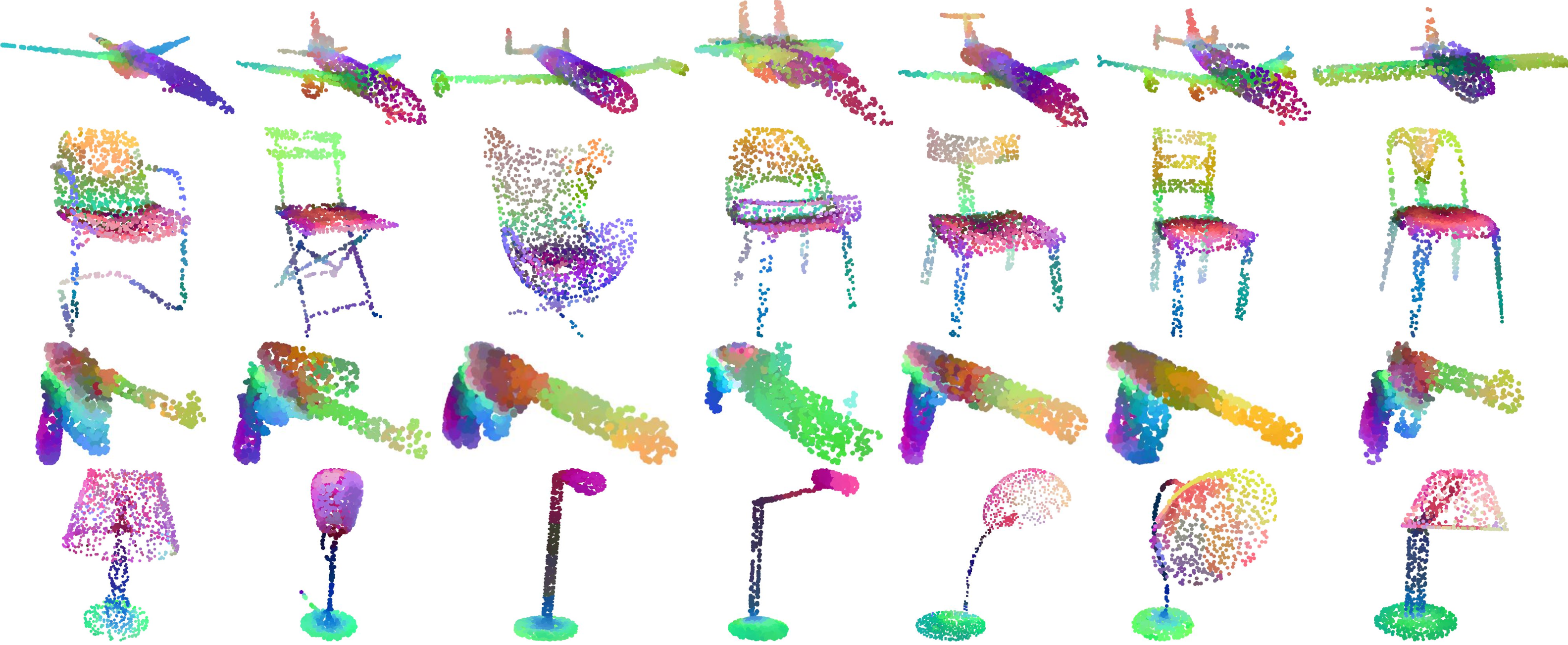}

   \caption{\textbf{TSNE visualization of learned embeddings.} For
     each shape category, we take a fixed number of shapes and extract
     point embeddings from \surfit. We run TSNE on each category
     separately to project the 128-D embeddings to 3D color
     space. Notice that points belonging to same semantic parts are
     colored similarly, which indicates the consistency of learned
     embeddings.}\label{fig:tsne}
\end{figure*}

\begin{table*}[]
  \centering
  \addtolength{\tabcolsep}{-2.5pt}
\begin{tabular}{@{}lcccccccc}
\toprule
Samples/cls.    & \textbf{k=1}      & \textbf{k=3}      & \textbf{k=5}     & \textbf{k=10}                   & \textbf{k=20}     & \textbf{k=50}      & \textbf{k=100}      & \textbf{k=max}   \\
\midrule
  Baseline      & 53.15 $\pm$ 2.4 & 59.54 $\pm$ 1.4  & 68.14 $\pm$ 0.9 & 71.32 $\pm$ 0.5  & 75.22 $\pm$ 0.8  &  78.79 $\pm$ 0.4  &  79.67 $\pm$ 0.3   & \textbf{81.40 $\pm$ 0.4}  \\

\hline
  \surfit           & \textbf{63.14 $\pm$ 3.4} & \textbf{71.24 $\pm$ 1.3}  & \textbf{73.75 $\pm$ 0.7}   & \textbf{75.03 $\pm$ 0.9} & \textbf{76.73 $\pm$ 0.5}    &  \textbf{79.28 $\pm$ 0.2}  &  \textbf{80.16 $\pm$ 0.2}  &    80.40 $\pm$ 0.1 \\ 

  \bottomrule
\end{tabular}
\addtolength{\tabcolsep}{2.5pt}

\caption{\textbf{Few-shot segmentation on the ShapeNet dataset} (\textit{class
    avg. IoU} over 5 rounds).  The number of shots or samples per
  class is denoted by $k$ for each of the 16 ShapeNet shape categories used
  for supervised training. Our proposed method \surfit
  consistently outperforms the baselines. 
  }\label{table:shapenet}
\end{table*}
\subsection{Datasets}
As a source of unlabeled data for the task of self supervision, we use
the ShapeNet Core dataset ~\cite{Chang2015ShapeNetAI}, which consists
of 55 categories with 55,447 meshes in total. We sample these meshes
uniformly to get $2048$ points per shape. For the task of few-shot
semantic segmentation, we use the ShapeNet Semantic Segmentation
dataset, which consists of 16,881 labeled point clouds across 16 shape
categories with total 50 part categories.

We also evaluate our method on the PartNet dataset
\cite{Mo_2019_CVPR}.  This dataset provides \textit{fine-grained}
semantic segmentation annotation for various 3D shape categories,
unlike the more coarse-level shape parts in the ShapeNet dataset. We
use 12 categories from ``level-3'', which denotes the finest level of
segmentation.

For training different approaches in few-shot framework, we remove test shapes of labeled dataset $\mathcal{X}_l$ from our self-supervision dataset $\mathcal{X}_u$. This avoids train-test set overlap.

\subsection{Few-shot part segmentation on ShapeNet }
For each category from the ShapeNet part 
segmentation dataset, we
randomly sample $k$ labeled shapes ($16\times k$ in total) and use
these for training the semantic segmentation component of our
model. We use the entire training set from the ShapeNet core dataset
for self supervision task. We train a \textit{single} model
on all 16 shape categories of ShapeNet.

\paragraph*{Baselines.} Our first baseline takes
the PointNet++ as the base architecture and trains it from scratch on
only labeled training examples, using $k$ labeled shapes per category
in a few-shot setup. Second, we create a reconstruction baseline -- we
use a PointNet++ as a shared feature extractor, which extracts a
global shape encoding that is input to an AtlasNet decoder
\cite{atlasnet} with 25 charts. A separate decoder is used to predict
per-point semantic labels. This network is trained using a Chamfer
distance-based reconstruction loss using the entire unlabeled training
set and using $k$ labeled training examples. 
We use 5 different rounds with sampled labeled sets at various
values of $k$ and report their average performance. We train a single model for
all categories.

\paragraph*{Discussion of results.}
\begin{table}[t]
\centering
\addtolength{\tabcolsep}{-2.5pt}
\begin{tabular}{@{}lcc|cc}
\toprule
 \multirow{2}{*}{\textbf{Method}} &  1\%         &  5\%          &  1\%               &  5\%\\
                                  & \textbf{ins IoU} & \textbf{ins IoU}  & \textbf{cls IoU} & \textbf{cls IoU}\\
\midrule
 SO-Net~\cite{li2018so}                      &  64.0             & 69.0  & -      & -\\
 PointCapsNet~\cite{zhao20193d}              &  67.0             & 70.0  & -      & -\\
 MortonNet~\cite{MortonNet}                  &  -                & 77.1  & -      & -\\
 Deformation~\cite{wang2020few} & 68.9 & - &66.2 & - \\
 JointSSL~\cite{jointssl}                    &  71.9             & 77.4  & -      & -\\
 Multi-task~\cite{hassani2019unsupervised}   &  68.2             & 77.7  & -      & 72.1\\
 ACD \cite{selfsupacd} $\dagger$                       &  75.1             & 78.6 & 74.6    & 77.5\\ 
\hline
\surfit w/ Atlas                            & 73.8               & 78.6 & 74.5    & 78.9\\
\surfit w/ Cuboid                           & 74.6               & 78.6 & 75.2  & 78.6\\
\surfit w/ Ellipsoid                        & \textbf{75.4}      & \textbf{78.7}  & \textbf{75.3}  & \textbf{79.0}\\
\bottomrule
\end{tabular}

\caption{\small{\textbf{Comparison with state-of-the-art few-shot part segmentation methods on ShapeNet.} Performance is evaluated using \textit{instance-averaged} and \textit{class-averages} IoU. {$\dagger$} - We re-ran the publicly-released code from ACD~\cite{selfsupacd} on our data splits, ensuring fair comparison.
}}\label{tab:sota_shapenet}
\end{table}

Table \ref{table:shapenet} shows results on few-shot segmentation at
different number of labeled examples.

Our method \surfit performs better than the supervised baseline showing the effectiveness of our method as 
semi-supervised approach. 

In Table \ref{tab:sota_shapenet} we compare our approach with previous
methods using instance IOU and class IOU \cite{qi2017pointnetpp}, using $1\%$ and
$5\%$ of the labeled training set to train different methods.  

Note that instance
IOU is highly influenced by the shape categories with large number of
testing shapes \eg Chair, Table.  Class IOU, on the other hand gives
equal importance to all categories, hence it is a more robust
evaluation metric. \surfit performs better than previous self-supervised \cite{MortonNet,selfsupacd,hassani2019unsupervised} and semi-supervised \cite{jointssl,wang2020few} approaches. Notable our approach significantly outperforms learned deformation based approach proposed by Wang \etal \cite{wang2020few} \footnote{We run their code on all 16 categories on 5 different random splits}. \surfit with ellipsoid as primitive
outperforms previous methods including \surfit baseline with AtlasNet. Interestingly, \surfit with AtlasNet outperforms all previous
baselines except ACD\footnote{We run their code on our random split for fair comparison.}. We speculate that since primitives predicted by AtlasNet are highly overlapped and less localized in comparison to our ellipsoid and cuboid fitting approaches as shown in Figure \ref{fig:primitives}, this results in worse performance of AtlasNet. 

\surfit with ellipsoid primitives gives the best results, outperforming ACD without having to rely on an external black-box method for the self-supervisory training signal. This shows that our approach of primitive fitting in an end-to-end trainable manner is better than training a network using contrastive learning guided by approximate 
convex decomposition of water-tight meshes as proposed by ACD.

In Figure \ref{fig:shapenet} we show predicted semantic labels using \surfit along with fitted ellipsoids. In Figure \ref{fig:primitives} we show outputs of various self
supervision techniques using primitive fitting. We experimented with
both cuboid and ellipsoid fitting as a self supervision task. We
observed that both performs similar qualitatively and
quantitatively. The fitted primitives using ellipsoid/cuboid fitting
approaches are more aligned with different parts of the shape in
comparison to the outputs of AtlasNet.
\addtolength{\tabcolsep}{2.5pt}
\begin{table}[]
\centering
\begin{tabular}{lcccc}
\toprule
size           & 0    & 2.5k  & 25k  & 52k\\
\hline
class avg. IOU & 68.1 & 68.9  & 72.6 & \textbf{73.7}\\
\bottomrule
\end{tabular}

\caption{Effect of the size of unlabeled dataset used for self-supervision on 5-shot semantic segmentation on ShapeNet.}\label{unsup}
\end{table}
\paragraph*{Effect of the size of unlabeled dataset.}
In the Table \ref{unsup} we show the effect of size of unlabeled dataset used
for self-supervision. We observe improvement in performance of 5-shot semantic
segmentation task with increase in unlabeled dataset.

\paragraph*{Effect of similarity loss.}
Similarity loss is only used in the initial stage of training as it prevents the network converging to a local minima at this early phase. Without this procedure, our performance is similar to Baseline (training from scratch). However, using only similarity loss (without reconstruction loss) leads to worse results (44.2 mIOU) than Baseline (68.1 mIOU) on few-shot k=5 setting.

\paragraph*{Analysis of learned point embeddings}
In Figure \ref{fig:nmi} we quantitatively analyze the performance of
clustering induced by \surfit and compare it with clustering
obtained by running the K-Means algorithm directly on point clouds. We take
$340$ shapes from the Airplane category of ShapeNet for this experiment and show the
histogram of Normalized Mutual Information (NMI) \cite{NMI} and the
precision-recall curve \cite{Fowlkes} between predicted clusters and ground truth part labels. Our approach produces clusters
with higher NMI ($54.3$ vs $35.4$), which shows better alignment of our predicted point clusters with the ground truth part labels. Our approach also produces higher precision clusters
in comparison to K-Means at equivalent recall, which shows the tendency
of our algorithm to over-segment a shape.  To further analyze the consistency
of learned point embedding across shapes, we use TSNE
\cite{vandermaaten14a} to visualize point embedding by projecting them
to 3D color space. We use a fixed number of shapes for each category and
run TSNE on each category separately. Figure \ref{fig:tsne} shows
points belonging to same semantic parts are consistently projected to
similar colors, further confirming the consistency of learned
embeddings.

\subsection{Few-shot segmantic segmentation on PartNet }
Here we experiment on the PartNet dataset for the task of few-shot
semantic segmentation. For each category from this
dataset, we randomly sample $k$ labeled shapes and use these for
training semantic segmentation part of the architecture. Similar to
our previous experiment, we use the complete training shapes from the
ShapeNet Core dataset for the self-supervision task. We choose DGCNN as a
backbone architecture for this experiment. Unlike our
ShapeNet experiments, in the PartNet experiment we train a separate model
for each category, as done in the original paper \cite{Mo_2019_CVPR}.

Similar to our previous experiment, we create two baselines -- 1) we
train a network from scratch providing only $k$ labeled examples, and
2) we train the AtlasNet on the entire unlabeled training set using the
self-supervised reconstruction loss and only $k$ labeled examples as supervision.

Table \ref{table:partnet} shows part avg.~IOU for the different
methods.  The AtlasNet method shows improvement over the purely-supervised baseline. \surfit shows
improvement over both AtlasNet and Baseline, indicating the effectiveness
of our approach in the \textit{fine-grained} semantic segmentation setting as well. We
further experiment with cuboids as the primitive, which achieves similar performance
as using ellipsoid primitives, consistent with our previous ShapeNet results.

\begin{table}[]
  \centering
    \addtolength{\tabcolsep}{-2.5pt}
\begin{tabular}{@{\extracolsep{1pt}}llll}
\toprule
Samples/cls.            & \textbf{k=10}      & \textbf{k=20}          & \textbf{k=40} \\
\midrule 
  Baseline              &  27.2$\pm$0.7  &  31.6$\pm$0.6              &     36.7$\pm$0.9          \\
\hline
  \surfit w/ Atlas     &  28.5$\pm$0.7  &  31.7$\pm$0.7               &     36.5$\pm$0.7         \\
\surfit w/ Cuboid                 &  29.3$\pm$0.4  &  32.4$\pm$0.7    &   37.5$\pm$0.5            \\
  \surfit w/ Ellipsoid             &  \textbf{29.4$\pm$0.7}  &  \textbf{32.6$\pm$ 0.6}  &   \textbf{37.6$\pm$0.4}            \\
\bottomrule
\end{tabular}
\addtolength{\tabcolsep}{2.5pt}

\caption{\textbf{Few-shot segmentation on the PartNet dataset} (\textit{part
    avg. IoU} over 5 rounds).  The number of shots or samples per
  class is denoted by $k$ for each of the 12 PartNet categories used
  for supervised training. Our proposed method \surfit
  consistently outperforms the baseline.}\label{table:partnet}

\end{table}

\paragraph*{Effect of intersection loss.} We also experiment with 
adding intersection loss while training \surfit with ellipsoid 
primitives, as shown in Table \ref{tab:intersection}. This gives 
improvements only on the PartNet dataset, and we speculate that since PartNet 
contains fine-grained segmentation of shapes, minimizing overlap 
between primitives here is more helpful than in the ShapeNet dataset, which 
contains only a coarse level of segmentation. On 
ShapeNet, we observed that the intersection loss is beneficial 
for 9 out of 16 categories (Airplane, Bag, Chair, Guitar, Lamp, Laptop, Mug, Rocket and Table). On the PartNet dataset we observed the intersection loss is beneficial for 8 out of 12 categories (Bed, Bottle, Dishwasher, Display, Knife, Microwave, StorageFurniture and TrashCan). In general, we observed that the benefits from this loss function varies with the category of shape and the coarseness of the segmentation. Similar observation was also made by Kawana \etal \cite{neuralstardomain}.

\begin{table}[]
\addtolength{\tabcolsep}{-5pt}
\begin{tabular}{l|cc|cc}
\toprule
\multirow{2}{*}{\textbf{Method}}     & \multicolumn{2}{c}{\begin{tabular}[c]{@{}c@{}}\textbf{ShapeNet}\\cls. IoU\end{tabular}} & \multicolumn{2}{|c}{\begin{tabular}[c]{@{}c@{}}\textbf{PartNet}\\ part avg. IoU\end{tabular}} \\
\cline{2-5}
                            & k=1\%                                    & k=5\%                                    & k=10                                   & k=20                                  \\
\hline
\surfit w/ Ellipsoids       & \textbf{75.3}                                   & \textbf{79.0}                                   & 28.9                                   & 32.1                                  \\
\surfit w/ Ellipsoids+inter & 75.0                                   & \textbf{79.0 }                                  & \textbf{29.4}                                   & \textbf{32.6}\\                                 
\bottomrule
\end{tabular}
\addtolength{\tabcolsep}{5pt}

\caption{\textbf{Effect of intersection loss.} Average performance over all categories for \surfit trained with intersection loss (+\textit{inter}) and without intersection loss on ShapeNet and PartNet dataset. \surfit trained with intersection loss (+\textit{inter}) gives improvement on PartNet dataset.}\label{tab:intersection}
\vspace{-7mm}
\end{table}

\section{Conclusion}
We  propose a simple  semi-supervised learning approach, \surfit, for learning point embeddings for few shot semantic segmentation. Our approach learns to decompose an unlabeled point cloud into a set of geometric primitives, such as ellipsoids and cuboids, or alternatively  deformed planes as in AtlasNet.
We provide an end-to-end trainable framework for incorporating this task into standard network architectures for point cloud segmentation. \surfit can be readily applied to existing architectures for semantic segmentation and shows improvements over fully-supervised baselines and other approaches on the ShapeNet and PartNet datasets. This indicates that learning to reconstruct a shape using primitives can induce representations useful for discriminative downstream tasks. \surfit also has limitations: we currently rely on basic primitives of a single type (\ie ellipsoid, cuboids, or deformed planes). Predicting combinations of primitives or other types of primitives for each shape could be useful to  capture more part variability. In addition, our primitives capture the shape rather coarsely, making our representations less fit for fine-grained tasks. 
Finally, it would be interesting to explore unsupervised approaches based on primitive and other forms of surface fitting.

\section{Acknowledgements}
The work is supported in part by NSF grants 1908669 and 1749833. The experiments were performed using equipment obtained under a grant from the Collaborative Fund managed by the Mass. Tech. Collaborative.
\bibliographystyle{eg-alpha-doi}  
\bibliography{egbib}

@article{hoffman1983parts,
  title={Parts of recognition},
  author={Hoffman, Donald D and Richards, Whitman},
  year={1983}
}

@inproceedings{NEURIPS2021_d1ee59e2,
 author = {Sun, Weiwei and Tagliasacchi, Andrea and Deng, Boyang and Sabour, Sara and Yazdani, Soroosh and Hinton, Geoffrey E and Yi, Kwang Moo},
 booktitle = {Advances in Neural Information Processing Systems},
%  editor = {M. Ranzato and A. Beygelzimer and Y. Dauphin and P.S. Liang and J. Wortman Vaughan},
%  pages = {24993--25005},
%  publisher = {Curran Associates, Inc.},
 title = {Canonical Capsules: Self-Supervised Capsules in Canonical Pose},
%  url = {https://proceedings.neurips.cc/paper/2021/file/d1ee59e20ad01cedc15f5118a7626099-Paper.pdf},
 volume = {34},
 year = {2021}
}

@article{generalizedcylinderLiu,
author = {Liu, Yanchao and Guo, Jianwei and Benes, Bedrich and Deussen, Oliver and Zhang, Xiaopeng and Huang, Hui},
title = {TreePartNet: Neural Decomposition of Point Clouds for 3D Tree Reconstruction},
year = {2021},
issue_date = {December 2021},
publisher = {Association for Computing Machinery},
address = {New York, NY, USA},
volume = {40},
number = {6},
% issn = {0730-0301},
% url = {https://doi.org/10.1145/3478513.3480486},
% doi = {10.1145/3478513.3480486},
abstract = {We present TreePartNet, a neural network aimed at reconstructing tree geometry from point clouds obtained by scanning real trees. Our key idea is to learn a natural neural decomposition exploiting the assumption that a tree comprises locally cylindrical shapes. In particular, reconstruction is a two-step process. First, two networks are used to detect priors from the point clouds. One detects semantic branching points, and the other network is trained to learn a cylindrical representation of the branches. In the second step, we apply a neural merging module to reduce the cylindrical representation to a final set of generalized cylinders combined by branches. We demonstrate results of reconstructing realistic tree geometry for a variety of input models and with varying input point quality, e.g., noise, outliers, and incompleteness. We evaluate our approach extensively by using data from both synthetic and real trees and comparing it with alternative methods.},
journal = {ACM Trans. Graph.},
month = {dec},
numpages = {16},
keywords = {optimization, procedural generation, 3D reconstruction, geometric modeling, deep learning, procedural modeling}
}

@article{primsurveryKaiser,
author = {Kaiser, Adrien and Ybanez Zepeda, Jose Alonso and Boubekeur, Tamy},
title = {A Survey of Simple Geometric Primitives Detection Methods for Captured 3D Data},
journal = {Computer Graphics Forum},
volume = {38},
number = {1},
% pages = {167-196},
keywords = {3D data, geometric primitives, shape analysis, shape abstraction, computational geometry, data fitting, I.3.5 Computing Methodologies/Computer Graphics: Computational Geometry and Object Modelling—Curve, surface, solid and object representations},
% doi = {https://doi.org/10.1111/cgf.13451},
% url = {https://onlinelibrary.wiley.com/doi/abs/10.1111/cgf.13451},
% eprint = {https://onlinelibrary.wiley.com/doi/pdf/10.1111/cgf.13451},
abstract = {Abstract The amount of captured 3D data is continuously increasing, with the democratization of consumer depth cameras, the development of modern multi-view stereo capture setups and the rise of single-view 3D capture based on machine learning. The analysis and representation of this ever growing volume of 3D data, often corrupted with acquisition noise and reconstruction artefacts, is a serious challenge at the frontier between computer graphics and computer vision. To that end, segmentation and optimization are crucial analysis components of the shape abstraction process, which can themselves be greatly simplified when performed on lightened geometric formats. In this survey, we review the algorithms which extract simple geometric primitives from raw dense 3D data. After giving an introduction to these techniques, from the acquisition modality to the underlying theoretical concepts, we propose an application-oriented characterization, designed to help select an appropriate method based on one's application needs and compare recent approaches. We conclude by giving hints for how to evaluate these methods and a set of research challenges to be explored.},
year = {2019}
}

@inproceedings{wang2020few,
  title={Few-Shot Learning of Part-Specific Probability Space for 3D Shape Segmentation},
  author={Wang, Lingjing and Li, Xiang and Fang, Yi},
  booktitle={Proceedings of the IEEE/CVF Conference on Computer Vision and Pattern Recognition},
  year={2020}
}

@inproceedings{binford1987bayesian,
  title={Bayesian inference in model-based machine vision},
  author={Binford, Thomas O and Levitt, Tod S and Mann, Wallace B},
  booktitle={Proceedings of the Third Conference on Uncertainty in Artificial Intelligence},
  year={1987}
 }

@inproceedings{binford1971visual,
  title={Visual perception by computer},
  author={Binford, I},
  booktitle={IEEE Conference of Systems and Control},
  year={1971}
}

@article{marr1978representation,
  title={Representation and recognition of the spatial organization of three-dimensional shapes},
  author={Marr, David and Nishihara, Herbert Keith},
  journal={Proceedings of the Royal Society of London. Series B. Biological Sciences},
  volume={200},
  number={1140},
  year={1978},
  publisher={The Royal Society London}
}

@INPROCEEDINGS{SharmaPEN, 
author={Sharma, Gopal and Kalogerakis, Evangelos and Maji, Subhransu},
booktitle={International Conference on 3D Vision (3DV)},
title={Learning Point Embeddings from Shape Repositories for Few-Shot Segmentation}, 
year={2019}
}

@article{biederman1987recognition,
  title={Recognition-by-components: a theory of human image understanding.},
  author={Biederman, Irving},
  journal={Psychological review},
  volume={94},
  number={2},
  year={1987},
  publisher={American Psychological Association}
}

@inproceedings{chen2019bae,
  title={{BAE-NET:} branched autoencoder for shape co-segmentation},
  author={Chen, Zhiqin and Yin, Kangxue and Fisher, Matthew and Chaudhuri, Siddhartha and Zhang, Hao},
  booktitle={Proceedings of the IEEE International Conference on Computer Vision},
%   pages={8490--8499},
  year={2019}
}

@inproceedings{li2018so,
  title={SO-Net: Self-organizing network for point cloud analysis},
  author={Li, Jiaxin and Chen, Ben M and Hee Lee, Gim},
  booktitle={Proceedings of the IEEE conference on computer vision and pattern recognition},
%   pages={9397--9406},
  year={2018}
}

@inproceedings{zhao20193d,
  title={{3D} point capsule networks},
  author={Zhao, Yongheng and Birdal, Tolga and Deng, Haowen and Tombari, Federico},
  booktitle={Proceedings of the IEEE Conference on Computer Vision and Pattern Recognition},
%   pages={1009--1018},
  year={2019}
}

@article{Yangcuboidfitting,
author = {Yang, Kaizhi and Chen, Xuejin},
title = {Unsupervised Learning for Cuboid Shape Abstraction via Joint Segmentation from Point Clouds},
year = {2021},
issue_date = {August 2021},
publisher = {Association for Computing Machinery},
address = {New York, NY, USA},
volume = {40},
number = {4},
% issn = {0730-0301},
% url = {https://doi.org/10.1145/3450626.3459873},
% doi = {10.1145/3450626.3459873},
abstract = {Representing complex 3D objects as simple geometric primitives, known as shape abstraction, is important for geometric modeling, structural analysis, and shape synthesis. In this paper, we propose an unsupervised shape abstraction method to map a point cloud into a compact cuboid representation. We jointly predict cuboid allocation as part segmentation and cuboid shapes and enforce the consistency between the segmentation and shape abstraction for self-learning. For the cuboid abstraction task, we transform the input point cloud into a set of parametric cuboids using a variational auto-encoder network. The segmentation network allocates each point into a cuboid considering the point-cuboid affinity. Without manual annotations of parts in point clouds, we design four novel losses to jointly supervise the two branches in terms of geometric similarity and cuboid compactness. We evaluate our method on multiple shape collections and demonstrate its superiority over existing shape abstraction methods. Moreover, based on our network architecture and learned representations, our approach supports various applications including structured shape generation, shape interpolation, and structural shape clustering.},
journal = {ACM Trans. Graph.},
month = {jul},
articleno = {152},
numpages = {11},
keywords = {3D structural representation, joint segmentation, 3D shape abstraction, point clouds}
}

@inproceedings{neuralstardomain,
author = {Kawana, Yuki and Mukuta, Yusuke and Harada, Tatsuya},
title = {Neural Star Domain as Primitive Representation},
year = {2020},
% isbn = {9781713829546},
% publisher = {Curran Associates Inc.},
% address = {Red Hook, NY, USA},
abstract = {Reconstructing 3D objects from 2D images is a fundamental task in computer vision, and an accurate structured reconstruction by parsimonious and semantic primitive representations has an even broader application range. When a target shape is reconstructed using multiple primitives, it is preferable that its fundamental properties, such as collective volume and surface, can be readily and comprehensively accessed so that the primitives can be treated as if they were a single shape. This becomes possible by a primitive representation with unified implicit and explicit representations. However, primitive representations in current approaches do not satisfy these requirements. To resolve this, we propose a novel primitive representation termed neural star domain (NSD) that learns primitive shapes in a star domain. We demonstrate that NSD is a universal approximator of the star domain; furthermore, it is not only parsimonious and semantic but also an implicit and explicit shape representation. The proposed approach outperforms existing methods in image reconstruction tasks in terms of semantic capability as well as sampling speed and quality for high-resolution meshes.},
booktitle = {Proceedings of the 34th International Conference on Neural Information Processing Systems},
articleno = {660},
% numpages = {12},
% location = {Vancouver, BC, Canada},
% series = {NIPS'20}
}

@inproceedings{hassani2019unsupervised,
  title={Unsupervised multi-task feature learning on point clouds},
  author={Hassani, Kaveh and Haley, Mike},
  booktitle={Proceedings of the IEEE International Conference on Computer Vision},
%   pages={8160--8171},
  year={2019}
}

@inproceedings{yang2018foldingnet,
  title={Foldingnet: Point cloud auto-encoder via deep grid deformation},
  author={Yang, Yaoqing and Feng, Chen and Shen, Yiru and Tian, Dong},
  booktitle={Proceedings of the IEEE Conference on Computer Vision and Pattern Recognition},
%   pages={206--215},
  year={2018}
}

@string{cvpr = {Proc. {CVPR}}}

@string{eccv = {Proc. {ECCV}}}

@string{iccv = {Proc. {ICCV}}}

@string{nips = {Proc. {NIPS}}}

@string{siggraph = {Proc. {SIGGRAPH}}}

@string{tog = {{ACM Trans. Graph.}}}

@inproceedings{qi2017pointnet,
  title={{PointNet}: Deep Learning on Point Sets for {3D} Classification and Segmentation},
  author={Qi, Charles R and Su, Hao and Mo, Kaichun and Guibas, Leonidas J},
  booktitle=cvpr,
  year={2017}
}

@inproceedings{qi2017pointnetpp,
  title={{PointNet++}: Deep Hierarchical Feature Learning on Point Sets in a Metric Space},
  author={Charles R. Qi and Li Yi and Hao Su and Leonidas Guibas},
  booktitle=nips,
  year={2017}
}

@inproceedings{kong2018recurrent,
  title={Recurrent pixel embedding for instance grouping},
  author={Kong, Shu and Fowlkes, Charless C},
  booktitle={Proceedings of the IEEE Conference on Computer Vision and Pattern Recognition},
%   pages={9018--9028},
  year={2018}
}

@article{Chang2015ShapeNetAI,
  title={ShapeNet: An Information-Rich 3D Model Repository},
  author={Angel X. Chang and Thomas A. Funkhouser and Leonidas J. Guibas and Pat Hanrahan and Qi-Xing Huang and Zimo Li and Silvio Savarese and Manolis Savva and Shuran Song and Hao Su and Jianxiong Xiao and Li Yi and Fisher Yu},
  journal={CoRR},
  year={2015},
  volume={abs/1512.03012}
}

@inproceedings{Muralikrishnan18,
 author = {Sanjeev Muralikrishnan and Vladimir G. Kim and Siddhartha Chaudhuri},
 title = {{Tags2Parts}: Discovering Semantic Regions from Shape Tags},
 booktitle = {Proc. CVPR},
 year = {2018},
 publisher = {IEEE},
}

@inproceedings{yang2019pointflow,
  title={Pointflow: 3d point cloud generation with continuous normalizing flows},
  author={Yang, Guandao and Huang, Xun and Hao, Zekun and Liu, Ming-Yu and Belongie, Serge and Hariharan, Bharath},
  booktitle={Proceedings of the IEEE International Conference on Computer Vision},
%   pages={4541--4550},
  year={2019}
}

@article{luo2020learning,
      title={Learning to Group: A Bottom-Up Framework for 3D Part Discovery in Unseen Categories},
      author={Luo, Tiange and Mo, Kaichun and Huang, Zhiao and Xu, Jiarui and Hu, Siyu and Wang, Liwei and Su, Hao},
      journal={arXiv preprint arXiv:2002.06478},
      year={2020}
}

@article{wang2019dynamic,
  title={Dynamic Graph CNN for Learning on Point Clouds},
  author={Wang, Yue and Sun, Yongbin and Liu, Ziwei and Sarma, Sanjay E and Bronstein, Michael M and Solomon, Justin M},
  journal={ACM Transactions on Graphics (TOG)},
  volume={38},
  number={5},
%   pages={1--12},
  year={2019},
  publisher={ACM New York, NY, USA}
}

@article{mzlsgm_abstraction_siga_09,
AUTHOR = "Ravish Mehra and Qingnan Zhou and Jeremy Long and Alla Sheffer and Amy Gooch and Niloy J. Mitra",
TITLE = "Abstraction of Man-Made Shapes",
JOURNAL = "{ACM} Transactions on Graphics",
VOLUME = "28",
NUMBER = "5",
% PAGES = "\#137, 1-10",
YEAR = "2009",
}

@article{gsmc_iwires_sig_09,
AUTHOR = "Ran Gal and Olga Sorkine and Niloy J. Mitra and  Daniel Cohen-Or",
TITLE = "iWIRES: An Analyze-and-Edit Approach to Shape Manipulation",
JOURNAL = "{ACM} Transactions on Graphics (Siggraph)",
VOLUME = "28",
NUMBER = "3",
% PAGES = "\#33, 1-10",
YEAR = "2009",
}

@article{Gori2017,
  author  = {Gori, Giorgio and Sheffer, Alla and Vining, Nicholas and Rosales, Enrique and Carr, Nathan and Ju, Tao},
  title   = {FlowRep: Descriptive Curve Networks for Free-Form Design Shapes},
  journal = {ACM Transaction on Graphics},
  year    = {2017},
  volume = {36},
  number = {4},
  publisher = {ACM},
  address = {New York, NY, USA}
}

@article{spheremesh,
		author = {Jean-Marc Thiery and Emilie Guy and Tamy Boubekeur},
		title = {Sphere-Meshes: Shape Approximation using Spherical Quadric Error Metrics},
		journal ={ACM Transaction on Graphics (Proc. SIGGRAPH Asia 2013)},
		year = {2013},
		volume = {32},
		number = {6},
		pages = {Art. No. 178},
}

@Article{obbcage,
author="Xian, Chuhua
and Lin, Hongwei
and Gao, Shuming",
title="Automatic cage generation by improved OBBs for mesh deformation",
journal="The Visual Computer",
year="2012",
volume="28",
number="1",
% pages="21--33",
}

@InProceedings{partnet,
author = {Mo, Kaichun and Zhu, Shilin and Chang, Angel X. and Yi, Li and Tripathi, Subarna and Guibas, Leonidas J. and Su, Hao},
title = {PartNet: A Large-Scale Benchmark for Fine-Grained and Hierarchical Part-Level 3D Object Understanding},
booktitle = {Computer Vision and Pattern Recognition (CVPR)},
year = {2019}
}

@inproceedings{atlasnet,
          title={{AtlasNet: A Papier-M\^ach\'e Approach to Learning 3D Surface Generation}},
          author={Groueix, Thibault and Fisher, Matthew and Kim, Vladimir G. and Russell, Bryan and Aubry, Mathieu},
          booktitle={Proceedings IEEE Conf. on Computer Vision and Pattern Recognition (CVPR)},
          year={2018}
        }

@inProceedings{mrt18,
  title={{Multiresolution Tree Networks for 3D Point Cloud Processing}},
  author = {Matheus Gadelha and Rui Wang and Subhransu Maji},
  booktitle={ECCV},
  year={2018}
}

@inproceedings{genova2019learning,
  title={Learning Shape Templates with Structured Implicit Functions},
  author={Genova, Kyle and Cole, Forrester and Vlasic, Daniel and Sarna, Aaron and Freeman, William T and Funkhouser, Thomas},
booktitle={International Conference on Computer Vision},
  year={2019}
}

@article{gdc,
 author = {Zhou, Yang and Yin, Kangxue and Huang, Hui and Zhang, Hao and Gong, Minglun and Cohen-Or, Daniel},
 title = {Generalized Cylinder Decomposition},
 journal = {ACM Trans. Graph.},
 volume = {34},
 number = {6},
 year = {2015},
 publisher = {ACM},
 address = {New York, NY, USA},
}

@article{MortonNet,
	author = {Thabet, Ali and Alwassel, Humam and Ghanem, Bernard},
	title = {{MortonNet: Self-Supervised Learning of Local Features in 3D Point Clouds}},
	journal = {arXiv},
	year = {2019},
	month = {Mar},
	eprint = {1904.00230},
% 	url = {https://arxiv.org/abs/1904.00230}
}

@inproceedings{deng2019cvxnet,
  title={CvxNet: Learnable Convex Decomposition},
  author={Boyang Deng and Kyle Genova and Soroosh Yazdani and Sofien Bouaziz and Geoffrey Hinton and Andrea Tagliasacchi},
  booktitle={Proceedings of the IEEE Conference on Computer Vision and Pattern Recognition},
  year={2020},
}

@misc{jointssl,
    title={Joint Supervised and Self-Supervised Learning for 3D Real-World Challenges},
    author={Antonio Alliegro and Davide Boscaini and Tatiana Tommasi},
    year={2020},
    journal={arXiv}
}

@inProceedings{selfsupacd,
  title={Label-Efficient Learning on Point Clouds using Approximate Convex Decompositions},
  author = {Matheus Gadelha and Aruni RoyChowdhury and Gopal Sharma and Evangelos Kalogerakis and Liangliang Cao and Erik Learned-Miller and Rui Wang and Subhransu Maji},
  booktitle={European Conference on Computer Vision (ECCV)},
  year={2020}
}

@inproceedings{carr2001reconstruction,
  title={Reconstruction and representation of 3D objects with radial basis functions},
  author={Carr, Jonathan C and Beatson, Richard K and Cherrie, Jon B and Mitchell, Tim J and Fright, W Richard and McCallum, Bruce C and Evans, Tim R},
  booktitle={Proceedings of the 28th annual conference on Computer graphics and interactive techniques},
%   pages={67--76},
  year={2001}
}

@inproceedings{macedo2011hermite,
  title={Hermite radial basis functions implicits},
  author={Macedo, Ives and Gois, Joao Paulo and Velho, Luiz},
  booktitle={Computer Graphics Forum},
  volume={30},
  number={1},
%   pages={27--42},
  year={2011},
  organization={Wiley Online Library}
}

@book{mve,
author = {Fang, Shu-Cherng and Puthenpura, Sarat},
title = {Linear Optimization and Extensions: Theory and Algorithms},
year = {1993},
isbn = {0139152652},
publisher = {Prentice-Hall, Inc.},
address = {USA}
}

@InProceedings{Mo_2019_CVPR,
    author = {Mo, Kaichun and Zhu, Shilin and Chang, Angel X. and Yi, Li and Tripathi, Subarna and Guibas, Leonidas J. and Su, Hao},
    title = {{PartNet}: A Large-Scale Benchmark for Fine-Grained and Hierarchical Part-Level {3D} Object Understanding},
    booktitle = {Computer Vision and Pattern Recognition (CVPR)},
    year = {2019}
}

@INPROCEEDINGS{Ionescu,
  author={C. {Ionescu} and O. {Vantzos} and C. {Sminchisescu}},
  booktitle={IEEE International Conference on Computer Vision (ICCV)}, 
  title={Matrix Backpropagation for Deep Networks with Structured Layers}, 
  year={2015},
  volume={},
  number={},
%   pages={2965-2973}
  }

@misc{Lingxiao,
  Author = {Lingxiao Li and Minhyuk Sung and Anastasia Dubrovina and Li Yi and Leonidas Guibas},
  Title = {Supervised Fitting of Geometric Primitives to 3D Point Clouds},
  Year = {2018},
  Eprint = {arXiv:1811.08988},
}

@inproceedings{Schulman2015,
author = {Schulman, John and Heess, Nicolas and Weber, Theophane and Abbeel, Pieter},
title = {Gradient Estimation Using Stochastic Computation Graphs},
year = {2015},
booktitle = {Neural Information Processing Systems},
}

@inproceedings{sharma2020parsenet,
      title={ParSeNet: A Parametric Surface Fitting Network for 3D Point Clouds}, 
      author={Gopal Sharma and Difan Liu and Subhransu Maji and Evangelos Kalogerakis and Siddhartha Chaudhuri and Radomír Měch},
      year={2020},
      booktitle={Computer Vision and Pattern Recognition (CVPR)}
}

@inproceedings{occnet,
    title={Occupancy Networks: Learning 3D Reconstruction in Function Space},
    author={Mescheder, Lars and Oechsle, Michael and Niemeyer, Michael and Nowozin, Sebastian and Geiger, Andreas},
    booktitle={Proceedings IEEE Conf. on Computer Vision and Pattern Recognition (CVPR)},
    year={2019}
}

@misc{ellipsoidsdf,
  title = {{Ellipsoid SDF}},
  author={Inigo Quilez},
  howpublished = {\url{https://www.iquilezles.org/www/articles/ellipsoids/ellipsoids.htm}},
  note = {Accessed: 2020-11-15}
}

@Inproceedings{superquadricsrevisited,
     title = {Superquadrics Revisited: Learning 3D Shape Parsing beyond Cuboids},
     author = {Paschalidou, Despoina and Ulusoy, Ali Osman and Geiger, Andreas},
     booktitle = {Computer Vision and Pattern Recognition (CVPR)},
     year = {2019}
}

@inproceedings{superquadricshierarchical,
    title = {Learning Unsupervised Hierarchical Part Decomposition of 3D Objects from a Single RGB Image},
    author = {Paschalidou, Despoina and Luc van Gool and Geiger, Andreas},
    booktitle = {Proceedings IEEE Conf. on Computer Vision and Pattern Recognition (CVPR)},
    year = {2020},
}

@inProceedings{shapehandles,
  title={Learning Generative Models of Shape Handles},
  author = {Matheus Gadelha and Giorgio Gori and Duygu Ceylan and Radomir Mech and Nathan Carr and Tamy Boubekeur and Rui Wang and Subhransu Maji},
  booktitle={IEEE Conference on Computer Vision and Pattern Recognition (CVPR)},
  year={2020}
}

@inProceedings{abstractionTulsiani17,
  title={Learning Shape Abstractions by Assembling Volumetric Primitives},
  author = {Shubham Tulsiani and Hao Su and Leonidas J. Guibas and Alexei A. Efros and Jitendra Malik},
  booktitle={Computer Vision and Pattern Regognition (CVPR)},
  year={2017}
}

@article{NMI,
author = {Strehl, Alexander and Ghosh, Joydeep},
title = {Cluster Ensembles --- a Knowledge Reuse Framework for Combining Multiple Partitions},
year = {2003},
issue_date = {3/1/2003},
publisher = {JMLR.org},
volume = {3},
% number = {null},
% issn = {1532-4435},
% url = {https://doi.org/10.1162/153244303321897735},
% doi = {10.1162/153244303321897735},
abstract = {This paper introduces the problem of combining multiple partitionings of a set of objects into a single consolidated clustering without accessing the features or algorithms that determined these partitionings. We first identify several application scenarios for the resultant 'knowledge reuse' framework that we call cluster ensembles. The cluster ensemble problem is then formalized as a combinatorial optimization problem in terms of shared mutual information. In addition to a direct maximization approach, we propose three effective and efficient techniques for obtaining high-quality combiners (consensus functions). The first combiner induces a similarity measure from the partitionings and then reclusters the objects. The second combiner is based on hypergraph partitioning. The third one collapses groups of clusters into meta-clusters which then compete for each object to determine the combined clustering. Due to the low computational costs of our techniques, it is quite feasible to use a supra-consensus function that evaluates all three approaches against the objective function and picks the best solution for a given situation. We evaluate the effectiveness of cluster ensembles in three qualitatively different application scenarios: (i) where the original clusters were formed based on non-identical sets of features, (ii) where the original clustering algorithms worked on non-identical sets of objects, and (iii) where a common data-set is used and the main purpose of combining multiple clusterings is to improve the quality and robustness of the solution. Promising results are obtained in all three situations for synthetic as well as real data-sets.},
journal = {J. Mach. Learn. Res.},
month = mar,
% pages = {583–617},
numpages = {35},
keywords = {clustering, unsupervised learning, cluster analysis, mutual information, partitioning, multi-learner systems, knowledge reuse, ensemble, consensus functions}
}

@article{Fowlkes,
author = { E. B.   Fowlkes  and  C. L.   Mallows },
title = {A Method for Comparing Two Hierarchical Clusterings},
journal = {Journal of the American Statistical Association},
volume = {78},
number = {383},
% pages = {553-569},
year  = {1983},
publisher = {Taylor & Francis}
}

@article{vandermaaten14a,
  author  = {Laurens van der Maaten},
  title   = {Accelerating t-SNE using Tree-Based Algorithms},
  journal = {Journal of Machine Learning Research},
  year    = {2014},
  volume  = {15},
  number  = {93},
%   pages   = {3221-3245},
%   url     = {http://jmlr.org/papers/v15/vandermaaten14a.html}
}

@article{li2018sonet,
      title={SO-Net: Self-Organizing Network for Point Cloud Analysis},
      author={Li, Jiaxin and Chen, Ben M and Lee, Gim Hee},
      journal={arXiv preprint arXiv:1803.04249},
      year={2018}
}

@inproceedings{PointContrast2020,
    author = {Xie, Saining and Gu, Jiatao and Guo, Demi and  Qi, Charles R. and Guibas, Leonidas and Litany and Or},
    title = {PointContrast: Unsupervised Pre-training for 3D Point Cloud Understanding},
    booktitle = {ECCV},
    year = {2020},
}

@article{efficientransac,
author = {Schnabel, R. and Wahl, R. and Klein, R.},
title = {Efficient RANSAC for Point-Cloud Shape Detection},
journal = {Computer Graphics Forum},
volume = {26},
number = {2},
% pages = {214-226},
keywords = {large point-clouds, geometry analysis, shape fitting, localized RANSAC, primitive shapes, I.4.8: Scene Analysis Shape; Surface Fitting, I.3.5: Computational Geometry and Object Modeling Curve, surface, solid, object representations},
% doi = {https://doi.org/10.1111/j.1467-8659.2007.01016.x},
% url = {https://onlinelibrary.wiley.com/doi/abs/10.1111/j.1467-8659.2007.01016.x},
% eprint = {https://onlinelibrary.wiley.com/doi/pdf/10.1111/j.1467-8659.2007.01016.x},
abstract = {Abstract In this paper we present an automatic algorithm to detect basic shapes in unorganized point clouds. The algorithm decomposes the point cloud into a concise, hybrid structure of inherent shapes and a set of remaining points. Each detected shape serves as a proxy for a set of corresponding points. Our method is based on random sampling and detects planes, spheres, cylinders, cones and tori. For models with surfaces composed of these basic shapes only, for example, CAD models, we automatically obtain a representation solely consisting of shape proxies. We demonstrate that the algorithm is robust even in the presence of many outliers and a high degree of noise. The proposed method scales well with respect to the size of the input point cloud and the number and size of the shapes within the data. Even point sets with several millions of samples are robustly decomposed within less than a minute. Moreover, the algorithm is conceptually simple and easy to implement. Application areas include measurement of physical parameters, scan registration, surface compression, hybrid rendering, shape classification, meshing, simplification, approximation and reverse engineering.},
year = {2007}
}

@article{SilvermanDensity,
author = {Läuter, H.},
title = {Silverman, B. W.: Density Estimation for Statistics and Data Analysis. Chapman \& Hall, London – New York 1986, 175 pp.},
journal = {Biometrical Journal},
volume = {30},
number = {7},
% pages = {876-877},
% doi = {https://doi.org/10.1002/bimj.4710300745},
% url = {https://onlinelibrary.wiley.com/doi/abs/10.1002/bimj.4710300745},
% eprint = {https://onlinelibrary.wiley.com/doi/pdf/10.1002/bimj.4710300745},
year = {1988}
}

@book{mardia2009directional,
  title={Directional statistics},
  author={Mardia, Kanti V and Jupp, Peter E},
  volume={494},
  year={2009},
  publisher={John Wiley \& Sons}
}

@book{magnus99,
  abstract = {This book provides a self-contained and unified treatment of matrix differential calculus, aimed at econometricians and statisticians. It can be used as a textbook for senior undergraduate or graduate courses on the subject, and will also be a valuable source for professional econometricians and statisticians who want to learn and apply these important techniques. The authors base their approach on differentials rather than derivatives, and they show that the use of differentials is elegant, easy and considerably more useful in applications. No specialist knowledge of matrix algebra or calculus is required, since the basics of matrix algebra are covered in the first three chapters with a thorough treatment of multivariable calculus provided in Chapters Four to Seven. Exercises are included in each chapter, and many examples which illustrate applications of the theory are considered in detail.},
  added-at = {2014-03-16T13:03:09.000+0100},
  author = {Magnus, Jan R. and Neudecker, Heinz},
  biburl = {https://www.bibsonomy.org/bibtex/2c82317ecaf30079fa28c0fd5774e6780/ytyoun},
  edition = {Second},
  interhash = {76bd08d7316ac86cb1a7b88078010d57},
  intrahash = {c82317ecaf30079fa28c0fd5774e6780},
%   isbn = {0471986321 9780471986324 047198633X 9780471986331},
  keywords = {calculus economics linear.algebra matrix textbook},
  publisher = {John Wiley},
%   refid = {40467399},
%   timestamp = {2016-05-31T14:11:33.000+0200},
  title = {Matrix Differential Calculus with Applications in Statistics and Econometrics},
  year = 1999
}

@String{tog = "ACM TOG"}
\appendix
\section{Supplementary Material}\label{sec:supp}
\paragraph{Additional training details}
For few-shot experiments on Shapenet dataset, we train PointNet++ architecture on 4 Nvidia 1080Ti GPUs. We use batch size of 24 for all methods. The K-shot baseline is trained for
approximately $K\times 200$ iterations while \surfit is always trained with approximately $K\times 1000$ iterations. The average time to predict primitives for a single shape is 98 ms on a single Nvidia 1080Ti GPU.

\begin{figure}[h!]
\includegraphics[]{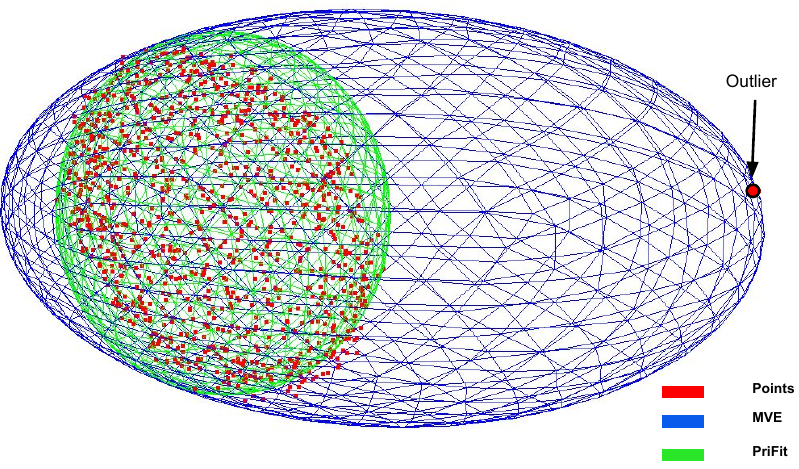}
\caption{\textbf{Robustness to outliers.} An example of outlier-robust fitting with our method in contrast to MVE (minimum volume ellipsoid) that is sensitive to outliers. Our fitting result shown in green closely fits the input points (red) while ignoring the outlier, whereas MVE approach (blue) is sensitive to the outlier.}\label{fig:mve}
\end{figure}

\paragraph{Robustness of ellipsoid fitting.}
Fig. \ref{fig:mve} shows the robustness of our approach to outliers in comparison to minimum volume ellipsoid (MVE)~\cite{mve}.   Our approach takes into account the membership of the point to a cluster. For this example, we use a simple membership function $1/r^{\frac{1}{4}}$, where r is the distance of point from the center of the cluster and incorporates these per-point weights to estimate the parameters of ellipsoid in a closed form using SVD.

\end{document}